%% file: main.tex
\documentclass[draftcls,onecolumn]{IEEEtran}

\usepackage{graphicx}
\usepackage{booktabs}
\usepackage[accsupp]{axessibility}
\usepackage{orcidlink}
\usepackage[table]{xcolor}
\usepackage{marvosym}
\usepackage{algorithm}
\usepackage{algpseudocode}
\usepackage{amsmath,amssymb}
\usepackage{amsthm}

\theoremstyle{definition}
\newtheorem{definition}{Definition}

\theoremstyle{remark}
\newtheorem{remark}{Remark}
\usepackage{cite}
\usepackage{url}
\usepackage{hyperref}

\usepackage[T1]{fontenc}
\usepackage[utf8]{inputenc}
\usepackage[section]{placeins}



\input{preamble}

\begin{document}

% Title

\title{Wavelet-Guided Semantic Signal Compensation
for Inversion-Free Image Editing}

% Authors

\author{
Anqi Tang\textsuperscript{*},
Wenhao Sun\textsuperscript{*},
and Zhaoqiang Liu\textsuperscript{\Letter}
\thanks{\textsuperscript{*} Equal contribution.}
\thanks{\textsuperscript{\Letter} Corresponding author.}
\thanks{The authors are with the School of Computer Science and Engineering, University of Electronic Science and Technology of China.}
\thanks{E-mail: anqitang@std.uestc.edu.cn, 202511081644@std.uestc.edu.cn, zqliu12@gmail.com}
\thanks{Project Page: \url{https://zoey-tang-33.github.io/wavelet-edit-project}}
}

\maketitle

\begin{abstract}

Text-guided image editing aims to modify visual content according to a target prompt while preserving the background. Recent inversion-free image editing frameworks such as FlowEdit have demonstrated strong editing capability without requiring inversion. Empirically, FlowEdit can achieve substantial semantic changes under appropriate hyperparameter settings. 
However, we observe that under certain global attribute shifts, the editing trajectory may not effectively move away from the source distribution in the early timesteps. Our analysis suggests that in the high-noise regime, the dominant manifold-seeking flow toward the data manifold can reduce the influence of the text-conditioned direction, leading to limited global modification while background structures remain only moderately preserved. Inspired by this observation, we propose an inversion-free, frequency-aware semantic compensation strategy that strengthens the effective signal in the early stage of generation, while maintaining structural consistency in the background. 
The proposed method improves global editing capacity without sacrificing background fidelity.
\end{abstract}

\section{Introduction}

Diffusion models~\cite{song2020score,song2020denoising,ho2020denoising} and rectified flows (RFs)~\cite{lipmanflow,liu2022rectified} have enabled high-fidelity image synthesis by progressively transforming noise into realistic images, achieving strong generation quality in practice. Building on their strong generative priors, diffusion and flow models have also become foundations for text-guided image editing, where the goal is to change semantics under textual control while preserving the structure of input. Traditional methods~\cite{pan2023effective,garibi2024renoise,wallace2023edict,zhang2024exact,han2024proxedit} typically rely on explicit semantic inversion to modify content while preserving spatial layouts. However, this process is plagued by reconstruction drift and high computational overhead. 

\begin{figure*}[!bp]
    \centering
    \includegraphics[width=0.85\textwidth]{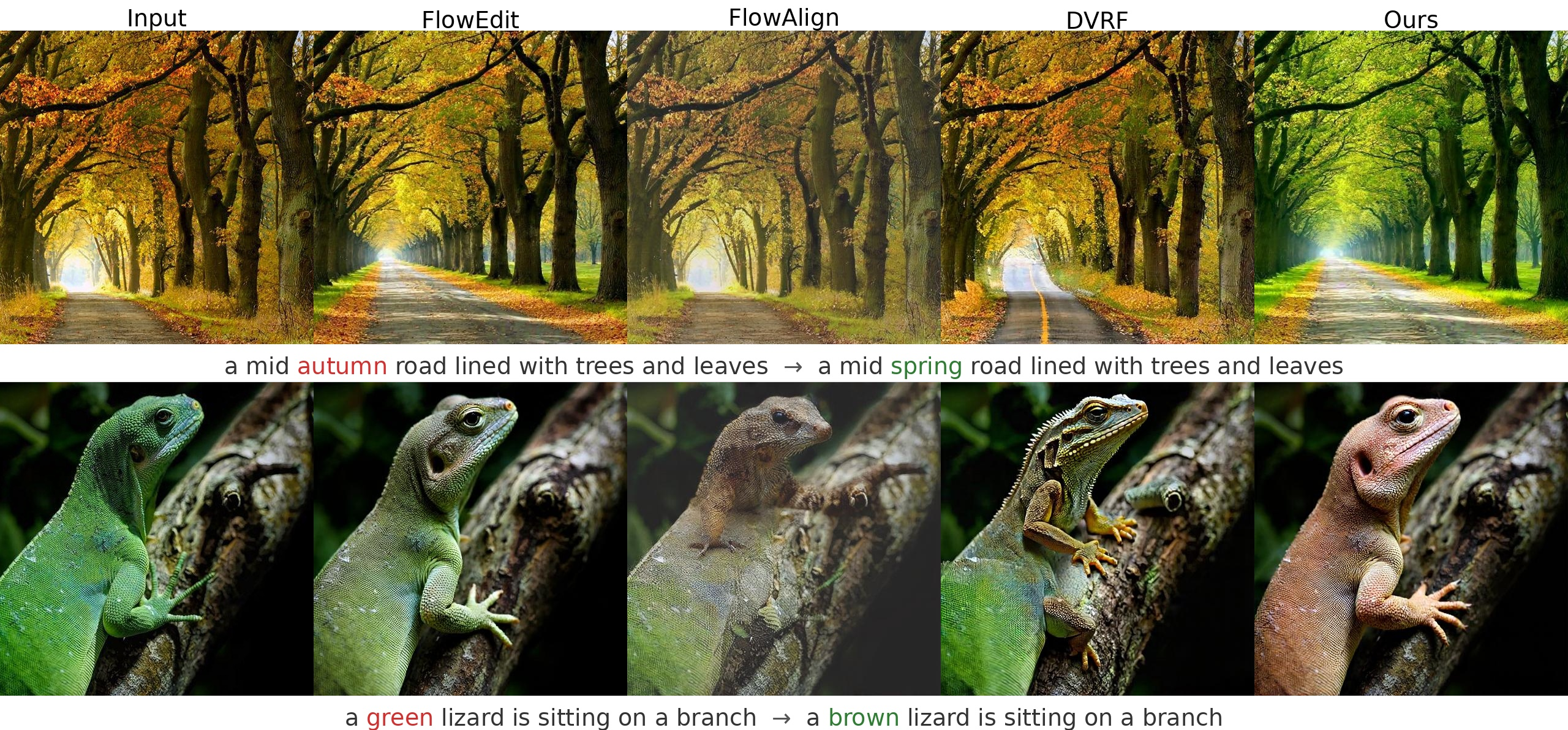}
    \caption{
    Failure cases of inversion-free RF-based editing methods, including FlowEdit~\cite{kulikov2024flowedit}, FlowAlign~\cite{kim2025flowalign}, and DVRF~\cite{beaudouin2025delta}, under global semantic modifications.
    When performing global attribute shifts, existing methods either introduce structural distortions, semantic drift, or incomplete color transformation.
    In contrast, our method achieves more consistent global attribute changes while preserving structural details.
    }
    \label{fig:global_shift_failure}
\end{figure*}

To avoid these bottlenecks, recent inversion-free editing frameworks~\cite{kulikov2024flowedit,mao2025tweezeedit,beaudouin2025delta,kim2025flowalign,li2025exact} modify images directly by dynamically estimating the velocity difference between a source trajectory and a target trajectory. By integrating this differential velocity field over time, these methods can guide the generative flow toward the target semantic domain without retracing the exact noise latent, thereby inherently retaining the structural priors of the source image.
While these inversion-free methods demonstrate impressive editing capability and structural preservation, we observe a clear limitation when they are applied to global semantic modifications, such as shifting overall color attributes, as illustrated in~\cref{fig:global_shift_failure}. We attribute this behavior to what can be described as semantic indistinguishability in the high-noise regime. At the initialization of the generation process, the latent state is dominated by unstructured noise. In this regime, the predicted velocity field is largely governed by the fundamental generative objective of transporting the noise distribution toward the coarse natural image manifold. This dominant manifold-seeking flow tends to outweigh the comparatively subtle semantic signal introduced by the text conditioning. As a result, the target trajectory remains strongly aligned with the structural prior inherited from the source image, limiting its ability to deviate toward new global attributes.
Consequently, during the early steps, when coarse structures and low-frequency components are established, the editing dynamics are primarily dictated by structural preservation rather than semantic transformation. By the time the noise level decreases and text conditioning becomes more influential, the global layout and color composition have already been largely fixed, significantly reducing the capacity for effective large-scale semantic modification.

To address this insufficient semantic deviation, it is necessary to extract a guidance signal that is disentangled from the dominant manifold-seeking flow. We propose same-point semantic probing, which evaluates both source and target conditions at the exact same latent state. By locking the spatial coordinates, the shared generative flow creates a baseline that allows us to obtain a co-located prompt-induced direction that is less confounded by the spatial separation between source and target latents. However, directly injecting this raw signal into the editing flow can introduce high-frequency perturbations that degrade the structural integrity of the source image.

To resolve this spatial-frequency conflict, we introduce a wavelet-based semantic signal compensation framework. Leveraging the 2D discrete Haar wavelet transform (DWT)~\cite{mallat2002theory}, we explicitly decouple the frequency bands of the semantic editing signal. We isolate the low-frequency components, while filtering out high-frequency detail coefficients. This prior is injected into the editing flow via a time-modulated update rule, ensuring maximum guidance during the early structural formation phase while quadratically decaying to preserve fidelity during the textural refinement phase.

The main contributions of this paper are summarized as follows:

\begin{itemize}
    \item We identify and analyze a limitation of inversion-free editing under rectified flow models, where the semantic editing signal becomes weak in high-noise regimes due to the dominance of manifold-seeking flow. We show that this effect limits the accumulation of directional deviation required for global attribute changes.

    \item We propose a wavelet-based semantic signal compensation mechanism that extracts low-frequency directions from identical latent states while suppressing high-frequency perturbations. This design strengthens global semantic modification while preserving structural consistency.

    \item  Extensive experiments, including quantitative evaluation, user studies, and frequency-domain analysis, demonstrate that our method achieves stronger and more consistent semantic editing across diverse tasks.
\end{itemize}

\section{Related Work}
\label{sec:rel_work}
Existing approaches in text-guided image editing can generally be categorized according to their generative backbones, namely diffusion models or rectified flows.
\subsection{Text-Guided Image Editing with Diffusion Models}

A principled family of methods first inverts the source image into the latent space of the diffusion model and then regenerates it under a modified text prompt. DDIM Inversion~\cite{hertz2022prompt,kim2022diffusionclip} retraces the deterministic sampling trajectory to estimate the noise latent that reconstructs the input, but the approximate nature of the inverse process often leads to reconstruction drift, especially under low-step settings.
To address this issue, subsequent works have proposed more consistent inversion schemes~\cite{pan2023effective,garibi2024renoise,wallace2023edict,zhang2024exact,han2024proxedit}, guidance-based manipulations~\cite{miyake2025negative,zhao2023null}, and higher-order solvers~\cite{hong2024exact,brack2024ledits++,feng2025dit4edit}.

There are also several works focusing on manipulating internal representations for structural preservation~\cite{hertz2022prompt,tumanyan2023plug,cao2023masactrl,xu2024inversion}. P2P~\cite{hertz2022prompt} controls cross-attention maps to align spatial layout, while PnP~\cite{tumanyan2023plug} and MasaCtrl~\cite{cao2023masactrl} inject spatial features and mutual self-attention queries, respectively, to maintain geometric consistency during editing. 
Beyond attention-level control, FreeDiff~\cite{wu2024freediff} exploits frequency-domain priors by employing a training-free progressive frequency truncation strategy to maintain structural consistency along the denoising steps.

While FreeDiff also leverages frequency decomposition, our method differs in operating domain and control objective. First, FreeDiff operates directly in the latent space via fast Fourier transform (FFT)~\cite{cooley1965algorithm}, aiming to constrain the generation by imposing the structural prior of source. This is a restrictive operation designed to prevent structure collapse in diffusion models. In contrast, our method operates in the velocity space of rectified flow models. Our goal is injection rather than constraint. We apply a wavelet transform to the prompt-induced editing signal and inject its low-frequency component into the velocity field to drive the necessary global deviations.

\subsection{Text-Guided Image Editing in Rectified Flows}

Within rectified flow models, inversion-based editing has been advanced by RF-Edit~\cite{wang2025taming}, 
which introduces a training-free high-order solver via Taylor expansion of the nonlinear ordinary differential equation term. 
RF-Inv~\cite{rout2025semantic} formulates semantic image inversion under rectified stochastic differential equations, 
enabling text-guided editing within the RF framework. 
FireFlow~\cite{deng2025fireflow} achieves a 3$\times$ speed-up with a second-order solver. 
ABM-Flow~\cite{ma2025adams} adopts a multistep predictor-corrector scheme with mask-guided latent injection. 
DNA-Edit~\cite{xie2025dnaedit} aligns noise trajectories directly to facilitate semantic manipulation in rectified flow models. 

A class of inversion-free methods bypasses explicit reconstruction altogether. FlowEdit~\cite{kulikov2024flowedit} formulates editing by aligning the velocity fields of the source and target trajectories. Building upon FlowEdit, recent works have focused on stabilizing the editing trajectory. FlowAlign~\cite{kim2025flowalign} and TweezeEdit~\cite{mao2025tweezeedit} assume shared noise conditions to simplify velocity-difference estimation under optimal control and path-regularization objectives. DVRF~\cite{beaudouin2025delta} generalizes these formulations with an additional correction term. 

While highly efficient, we observe that the editing signal of FlowEdit often diminishes in the high-noise regime due to the dominant manifold-seeking transport, leading to limited global semantic modifications. To address this semantic indistinguishability, our method introduces a wavelet-guided compensation mechanism to enforce stronger global semantic guidance during the early sampling steps.

In the broader context of image editing, frequency-domain constraints have been introduced to disentangle semantic changes from structural preservation. Recent inversion-free editing methods incorporate frequency-aware mechanisms, including FFT-based frequency stochasticity injection and frequency interactive attention, as well as wavelet-based multi-scale decomposition frameworks~\cite{teng2024fsi,liu2024fia,zhao2024wedit}. Different from them, our method retains the original pre-trained flow model without attention modifications.

\section{Preliminaries}

In this section, we summarize the formulation of rectified flow, and then describe the translation-coupled editing 
mechanism used in FlowEdit. 
For brevity, we provide a concise overview, and defer the full 
derivations and detailed discussions to Appendix~\ref{sec:prem_appen} . 
\subsection{Rectified Flow} 
Generative flow models aim to construct a transport between two distributions, a simple source distribution $p_1$ (typically standard Gaussian noise) to the complex data distribution $p_0$~\cite{lipmanflow}.Throughout this work, we follow the formulation in FlowEdit~\cite{kulikov2024flowedit},
where $t=1$ corresponds to pure noise and $t=0$ corresponds to clean data. The generative process is defined by an ordinary differential equation (ODE) flowing from $t=1$ to $t=0$, governed by a velocity field $\boldsymbol{v}_t$ with $\frac{\mathrm{d}\boldsymbol{x}_t}{\mathrm{d}t} = \boldsymbol{v}_t(\boldsymbol{x}_t)$. 

Rectified flow~\cite{liu2022rectified} adopts a linear interpolation path between noise and data with $\boldsymbol{x}_t = t\boldsymbol{x}_1 + (1-t)\boldsymbol{x}_0$. We discretize the time interval $[0,1]$ into $N$ subintervals with timesteps $1 = t_0 > t_1 > \dots > t_N = 0$. Starting from initial noise $\boldsymbol{x}_{t_0} \sim p_1$, we apply a first-order Euler discretization to numerically integrate the ODE:
\begin{equation}
\boldsymbol{x}_{t_{i+1}} = \boldsymbol{x}_{t_i} + (t_{i+1} - t_i)\boldsymbol{v}_{\btheta} (\boldsymbol{x}_{t_i}, t_i), \quad i \in \{0, \dots, N-1\},
\end{equation}
where $\boldsymbol{v}_{\btheta}$ denotes the pre-trained rectified flow model.

\subsection{FlowEdit}
FlowEdit~\cite{kulikov2024flowedit} formulates image editing by 
constructing a direct path from the source distribution 
to the target distribution via rectified flow. Given a source image \(\boldsymbol{x}^{\mathrm{src}}\), for any time step \(t \in [0, 1]\), we independently sample \(\boldsymbol{\epsilon} \sim \mathcal{N}(\boldsymbol{0}, \boldsymbol{I})\) and define $\boldsymbol{x}^{\mathrm{src}}_{t} = (1 - t)\boldsymbol{x}^{\mathrm{src}} + t \boldsymbol{\epsilon}$.
The idea of FlowEdit is to define the target trajectory $\boldsymbol{x}^{\mathrm{tar}}_{t}$ as a deviation from the source trajectory, driven by the semantic difference. Let $\boldsymbol{e}_{\mathrm{src}}$ and $\boldsymbol{e}_{\mathrm{tar}}$ denote the text embeddings of the source and target prompts, respectively. We define the delta velocity $\Delta \boldsymbol{v}_{t}$ between the target and source vector fields as 
\begin{equation}
\Delta \boldsymbol{v}_{t} \triangleq \boldsymbol{v}_\theta(\boldsymbol{x}^{\mathrm{tar}}_{t}, t; \boldsymbol{e}_{\mathrm{tar}}) - \boldsymbol{v}_\theta(\boldsymbol{x}^{\mathrm{src}}_{t}, t; \boldsymbol{e}_{\mathrm{src}}).
\end{equation}
FlowEdit constructs a target flow that preserves the structural dynamics of the source flow while integrating this semantic velocity. This corresponds to the coupled differential equation:
\begin{equation}
\mathrm{d}\boldsymbol{x}^{\mathrm{tar}}_t = \mathrm{d}\boldsymbol{x}^{\mathrm{src}}_t + \Delta \boldsymbol{v}_t \mathrm{d}t.
\label{eq:coupled_ode}
\end{equation}
By discretizing Eq.~\eqref{eq:coupled_ode}, we obtain the practical Euler update rule that combines the exact source transport with the semantic edit:
\begin{equation}
\boldsymbol{x}^{\mathrm{tar}}_{t_{i+1}} = \boldsymbol{x}^{\mathrm{tar}}_{t_i} + \underbrace{(\boldsymbol{x}^{\mathrm{src}}_{t_{i+1}} - \boldsymbol{x}^{\mathrm{src}}_{t_i})}_{\text{source transport}} + \underbrace{(t_{i+1} - t_i)\Delta \boldsymbol{v}_{t_i}}_{\text{semantic edit}}.
\end{equation}
This formulation ensures that the target generation faithfully follows the structural layout of the source image through the transport term, while $\Delta \boldsymbol{v}_{t_i}$ injects the necessary semantic alterations.

\subsection{Two-Dimensional Haar Wavelet Transform}
\label{sec:haar}

Given a spatial tensor $X \in \RR^{H \times W}$, the single-level 2D Haar transform~\cite{mallat2002theory}
decomposes each non-overlapping $2 \times 2$ block into four sub-bands
$(cA, cH, cV, cD)$ representing local averages and directional differences.
The forward transform is defined as
\begin{align}
    cA &= \tfrac{1}{2}(a+b+c+d), \\
    cH &= \tfrac{1}{2}(a-b+c-d), \\
    cV &= \tfrac{1}{2}(a+b-c-d), \\
    cD &= \tfrac{1}{2}(a-b-c+d),
\end{align}
and the inverse transform exactly reconstructs the original samples.
The transform is applied independently to each channel. Using the 2D Haar wavelet transform, we write the forward and inverse operators as $\DWT(\cdot)$ and $\IDWT(\cdot)$. The $cA$ sub-band encodes the local spatial average, while $cH$, $cV$, and $cD$ encode horizontal,
vertical, and diagonal local differences, respectively. The single-level inverse transform
reconstructs the original samples from these coefficients.

\section{Method}
\label{sec:method}
\begin{figure*}[t]
\centering
\includegraphics[width=0.9\textwidth]{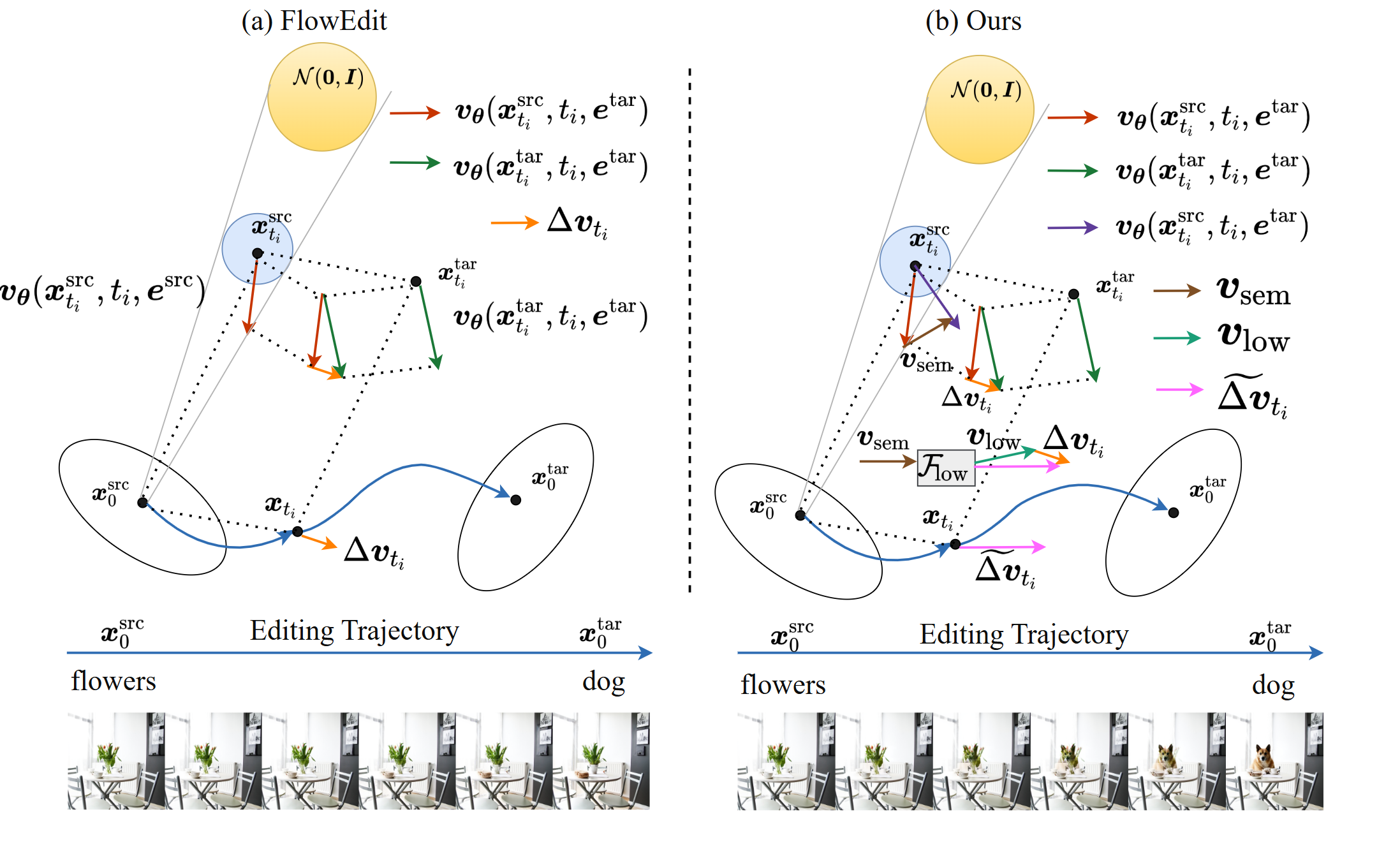}
\caption{ 
Illustration of trajectory editing with FlowEdit and our method.
FlowEdit directly applies a single residual direction $\Delta \bm{v}_{t_i}$ derived from the source and target velocity fields. Our method first estimates a co-located prompt-induced editing direction
($\bm{v}_{\mathrm{sem}}$), extracts its spatially coherent low-frequency
component ($\bm{v}_{\mathrm{low}}$) as an auxiliary global compensation, and
combines it with the original geometric update to obtain
$\widetilde{\Delta \bm{v}}_{t_i}$. The bottom row visualizes editing trajectory, where the scene gradually replaces the flowers with a dog while preserving the background layout.
}

\label{fig:method_comparison}
\end{figure*}

%% ================================================================
We present a training-free augmentation to FlowEdit~\cite{kulikov2024flowedit} 
that addresses its limited effectiveness in global attribute modification, 
such as large color shifts, as illustrated in~\cref{fig:global_shift_failure}. The overall mechanism is illustrated in~\cref{fig:method_comparison}.

\subsection{Motivation: Semantic Indistinguishability in the High-Noise Regime}
\label{sec:motivation}

Although FlowEdit produces visually compelling edits in many settings, we identify a failure mode that limits its capability for global semantic modifications. Specifically, while the geometric editing signal $\Delta \bv_{{t_i}}$ is non-zero during the high-noise phase in practice, it provides only weak and unreliable semantic directionality, which is often insufficient to initiate significant global attribute changes.

Since the early steps determine the global layout and dominant attributes of the generated image, the lack of informative guidance in this regime constrains global semantic modification.
Consequently, the target latent $\bx^{\text{tar}}_{t}$ accumulates insufficient directional momentum along the probability flow trajectory, yielding only weak global changes (e.g., uniform color shifts).
Instead, it remains closely aligned with the source trajectory and exhibits only minor attribute deviations.
When the noise level finally becomes low enough for the prompt to exert stronger semantic control, the global structure and color composition have already been fixed, making the edit ineffective.

\begin{figure*}[t]
    \centering
    \includegraphics[width=0.9\linewidth]{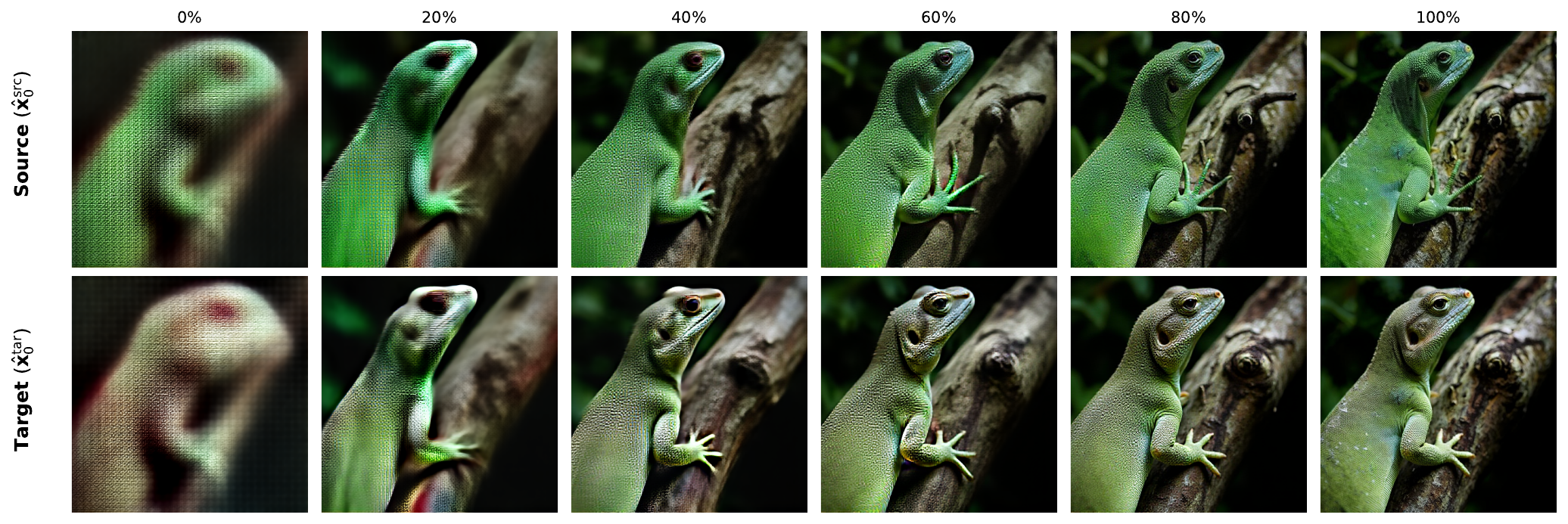}
    \caption{
    \textbf{Visualizing Semantic Signal Indistinguishability.}
    \textbf{Top Row:} Source trajectory.
    \textbf{Bottom Row:} Target trajectory conditioned on a prompt requesting a global attribute change (green $\rightarrow$ brown).
    }
    \label{fig:semantic_collapse}
\end{figure*}

Analogous to the Tweedie's formula in diffusion models, we recover the instantaneous expectation of the clean data $\hat{\bx}_0$ by projecting the flow forward.
Under the linear-trajectory assumption of rectified flow, the one-step estimator is as follows~\cite{martin2025pnpflow}:
\begin{equation} \label{eq:tweedie_est}
    \hat{\bx}_0(\bx_{t_i}) = \bx_{t_i} - t_i \cdot \bv_{\btheta}(\bx_{t_i}, t_i).
\end{equation}
Eq.~\eqref{eq:tweedie_est} provides a deterministic preview of the final structure from any intermediate state $t_i$.
As shown in~\cref{fig:semantic_collapse}, we examine $\hat{\bx}_0$ computed along the target trajectory of FlowEdit.
Despite the target prompt specifying a distinct attribute (e.g., a brown lizard), the semantic shift is largely suppressed by the source bias: Only faint target components appear, and the trajectory ultimately returns to the source appearance, causing the edit to fail.

\subsection{Problem Formulation}
\label{sec:problem}

Building on the analysis above, we formalize the main quantities involved
and describe the principles guiding the proposed method.

\begin{definition}[Semantic Editing Signal]
\label{def:vsem}
Given a noisy latent $\bm{x}$ at step $i$, source prompt embedding $\esrc$, and target prompt
embedding $\etar$, the semantic editing signal is defined as
\begin{equation}
    \vsem(\bm{x},\,{t_i})
    \;\triangleq\;
    \vtheta(\bm{x},\,{t_i};\,\etar) - \vtheta(\bm{x},\,{t_i};\,\esrc).
    \label{eq:vsem_def}
\end{equation}
\end{definition}

\begin{remark}
\label{rem:comparison}
The geometric editing signal $\Dv_{t_i}$ evaluates the velocity network at two distinct latent points $\bm{x}_t^{\mathrm{tar}}$ and $\bm{x}_t^{\mathrm{src}}$ under different prompts, so its magnitude can reflect both the prompt-conditioning difference and the spatial separation between the two inputs. By contrast, $\vsem$ evaluates both prompts at the same latent point, thereby removing the geometric confound caused by distinct input locations. We use this co-located prompt-induced direction as an auxiliary editing signal, rather than as an assumption that all semantic changes are represented by a particular frequency band.
\end{remark}

\begin{definition}[Low-Frequency Extraction Operator]
\label{def:lf}
Let $L$ denote the number of Haar wavelet decomposition levels. The \emph{low-frequency
extraction operator}
$\LF : \RR^{B \times C \times H \times W} \to \RR^{B \times C \times H \times W}$
is defined as the composition of an $L$-level forward two-dimensional Haar decomposition,
zeroing of all detail sub-bands at every level, and $L$-level inverse reconstruction:
\begin{equation}
    \LF(X)
    = \IDWT^{(1:L)}\!\!\left(
        cA^{(L)},\;
        \bigl\{\bm{0},\bm{0},\bm{0}\bigr\}_{l=1}^{L}
      \right),
    \label{eq:lf_op}
\end{equation}
where $cA^{(L)}$ is the $L$-th level approximation coefficient obtained by iterating the
single-level forward transform on successive approximation sub-bands.
\end{definition}
%\IDWT  \DWT

\subsection{Wavelet-Based Semantic Signal Compensation}
\label{sec:wssc}

\subsubsection{Multi-Level Decomposition and Low-Frequency Extraction}
\label{sec:multilevel}
Building upon the single-level Haar transform introduced in Sec.~\ref{sec:haar},
we construct a multi-level decomposition to isolate progressively coarser
structural components of the trajectory. The $L$-level forward decomposition iterates the single-level transform on the approximation
sub-band only. Setting $cA^{(0)} = X$, the recursion for $l = 1,\ldots,L$ is
\begin{equation}
    \bigl(cA^{(l)},\;cH^{(l)},\;cV^{(l)},\;cD^{(l)}\bigr)
    = \DWT\!\bigl(cA^{(l-1)}\bigr).
    \label{eq:fwd_multilevel}
\end{equation}
The operator $\LF$ defined in Definition~\ref{def:lf}
can be equivalently interpreted as a cascaded inverse reconstruction
starting from $cA^{(L)}$, where all detail sub-bands are set to zero
at each level.

\subsubsection{Semantic Editing Signal}
\label{sec:probe}

After computing the noise-averaged geometric signal
$\Dv_{t_i}$,
one additional network evaluation is performed at the source noisy latent $\xsrci$
under the target prompt. The semantic editing signal is defined as
\begin{equation}
    \vsem(\xsrci,\,t_i)
    = \vtheta\!\left(\xsrci,\,t_i;\,\etar\right)
      - \vtheta\!\left(\xsrci,\,t_i;\,\esrc\right).
    \label{eq:vsem_compute}
\end{equation}
Since both evaluations share the same input latent $\xsrci$, the difference $\vsem$ captures the co-located directional response induced by the change in text conditioning, while avoiding the confound introduced by evaluating the two prompts at distinct latent points. As the source-prompt velocity $\vtheta(\xsrci,t_i;\esrc)$ is already available, this additional evaluation requires only one extra forward pass per active timestep under the target prompt at $\xsrci$. This design choice distinguishes our approach from FlowAlign~\cite{kim2025flowalign} and CVC~\cite{li2025exact}. While they also utilize an additional network inference, they apply this correction to the target latent $\xtari$. In contrast, our method anchors the additional evaluation at the source latent $\xsrci$, so that the resulting signal serves as a prompt-conditioned auxiliary direction rather than a direct rectification of the edited trajectory.
\subsubsection{Frequency-Selective Injection and the Modified Update Rule}
\label{sec:injection}

While $\vsem$ provides a useful prompt-induced auxiliary direction, its raw form may also contain spatially incoherent perturbations caused by the random forward noise embedded in $\xsrci$. Directly injecting the full signal can therefore disturb fine structures in the output. Accordingly, the low-frequency extraction operator is applied to obtain
\begin{equation}
    \vlow = \LF\!\left(\vsem(\xsrci,\,t_i)\right),
    \label{eq:vlow}
\end{equation}
which retains the spatially slowly varying component of $\vsem$. This component is used only as a complementary compensation for globally coherent changes, such as overall color distribution, material appearance, and coarse spatial organization, which are often difficult to initiate during the high-noise phase. Notably, we do not assume that all semantic attributes are low-frequency, nor do we low-pass the entire editing update. The original geometric signal $\Dv_{t_i}$ remains active in Eq.~\eqref{eq:modified_ode}, allowing high-frequency structures, textures, and fine-grained attributes to still be handled by the standard editing dynamics.

The injection is modulated by the time-dependent weight $w(t_i) = \lambda\,t_i^2$, where 
$\lambda \geq 0$ controls the overall strength of the injected signal across timesteps. This quadratic schedule satisfies four
desirable properties: 1) At $t_i \approx 1$, the weight approximates $\lambda$,
maximally activating the compensation precisely where $\Dv_t$ is least informative. 2) At $t_i \approx 0$, the weight vanishes, reducing the ODE to the original FlowEdit update and preserving fine-detail fidelity. 3) The schedule is smooth and monotone, requiring no explicit threshold or switching criterion. 4) Setting $\lambda = 0$ recovers the original FlowEdit ODE exactly for all $t_i$. The complete modified update is
\begin{equation}
    \bm{x}_{t_{i+1}}
        = \bm{x}_{t_i}
          + (t_{i+1} - t_i)
            \!\left(
                \underbrace{\Dv_{t_i}}_{\text{geometric signal}}
                \;+\;
                \underbrace{
                    \lambda\,t_i^2
                    \cdot
                    \LF\!\bigl(\vsem(\xsrci,\,t_i)\bigr)
                }_{\text{global prompt compensation}}
            \right).
    \label{eq:modified_ode}
\end{equation}
The geometric signal $\Dv_{t_i}$ tracks the evolving discrepancy between the coupled noisy latents as the edited trajectory diverges from $\xsrc$, while the low-frequency auxiliary term provides a complementary global prompt-conditioning bias, especially in the high-noise regime where global attributes are difficult to establish.

For editing tasks confined to a specific spatial region, an optional binary mask
$M$ may be supplied. The masked combined editing
signal ${\widetilde{\Dv}}_{t_i}$ is
\begin{equation}
{\widetilde{\Dv}}_{t_i}
    = M \odot
      \Bigl(
          \Dv_{t_i}
          + \lambda\,t_i^2\,\vlow
      \Bigr),
    \label{eq:masked_signal}
\end{equation}
where $\odot$ denotes element-wise multiplication in the spatial domain.

\subsection{Frequency-Domain Analysis}
\label{sec:freq_analysis}
\begin{figure}[h]
  \centering
  \includegraphics[width=0.9\linewidth]{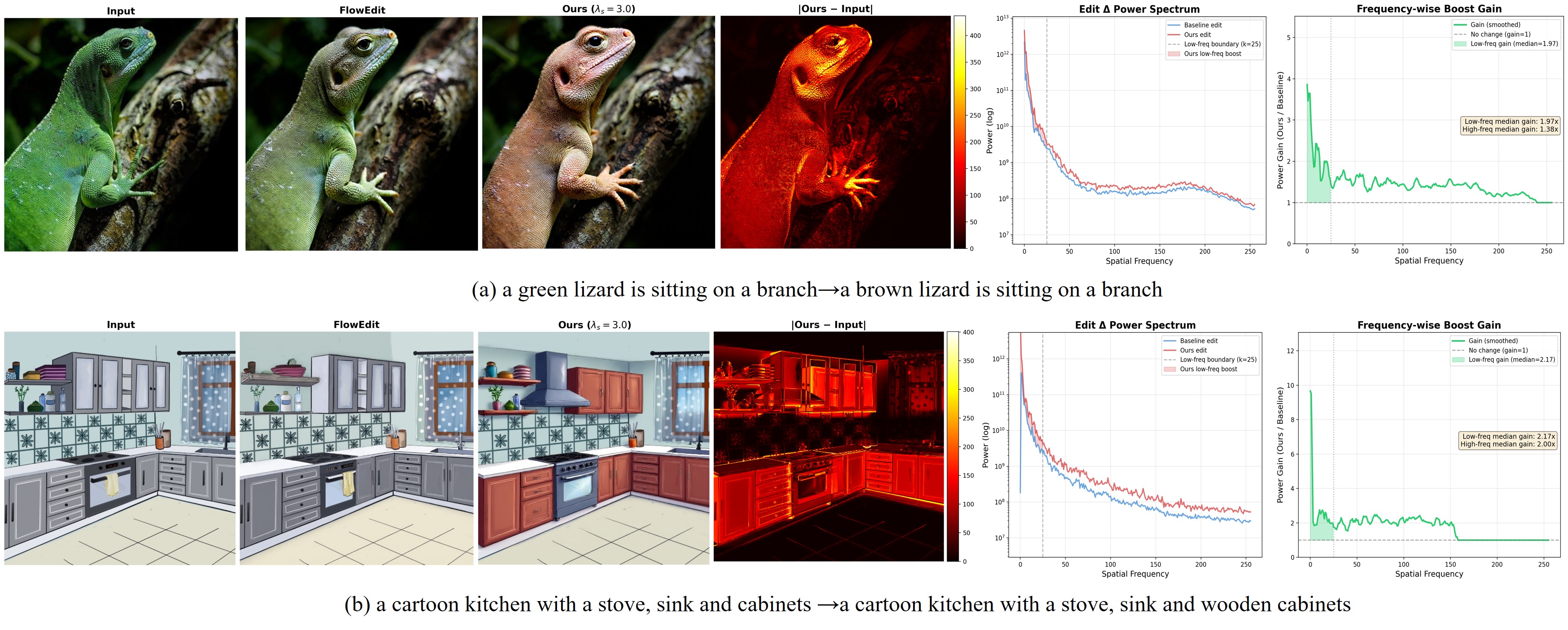}
  \caption{\textbf{Frequency-domain comparison of the editing delta.}
  Left: Qualitative editing examples.
  Middle: Radially averaged power spectral density (PSD) of the pixel-domain
  edit delta for FlowEdit (blue) and ours (red). The gray dashed line indicates
  the low-frequency boundary ($k=25$).
  Right: Frequency-wise power gain (Ours/Baseline).
 }
  \label{fig:freq_analysis}
\end{figure}
To examine the frequency-selective behavior of the proposed compensation, we analyze the spectral characteristics of the pixel-domain editing signal. Given a source image $\bm{x}^{\mathrm{src}}$, we compute the edit delta $\delta = \bm{x}_{t_N} - \bm{x}^{\mathrm{src}}$ for both FlowEdit and our method. Each delta map is converted to grayscale, transformed using a 2D discrete Fourier transform, and radially averaged to obtain the power spectral density (PSD) as a function of spatial frequency $k$. This analysis is intended to characterize how the compensation redistributes edit energy across frequency bands, rather than to serve as a standalone proof of semantic correctness. As shown in~\cref{fig:freq_analysis}, our method consistently produces higher spectral energy than the baseline in the low-frequency range, while remaining close to the baseline in the high-frequency regime. The frequency-wise gain curve further confirms this trend: The average gain is larger for $k < 25$, whereas only moderate amplification is observed at higher frequencies. These results support the intended behavior of the wavelet-guided compensation: It strengthens spatially coherent global changes without suppressing the high-frequency pathways. Therefore, the frequency-domain analysis should be interpreted as evidence for selective global compensation, complementing the visual and quantitative editing results.

\subsection{Pipeline of the Algorithm}
\label{sec:alg}

Algorithm~\ref{alg:main} presents the complete procedure.

\begin{algorithm}[!h]
\caption{Wavelet-Guided Semantic Signal Compensation
        Framework}
\label{alg:main}
\begin{algorithmic}[1]
\Require
    Pre-trained rectified flow model $\vtheta$,
    source latent $\xsrc$,
    source prompt embedding $\esrc$,
    target prompt embedding $\etar$,
    timestep schedule $\{t_i\}_{i=0}^{N}$,
    mask $M$,
    editing window $[n_{\mathrm{min}},\,n_{\mathrm{max}}]$,
    semantic strength $\lambda \geq 0$,
    wavelet levels $L$, the operator $\LF$
\Ensure Edited latent $\bm{x}_{t_N}$

\State Initialize $\bm{x}_{t_0} \leftarrow \xsrc$
\For{$i = N - n_{\mathrm{max}},\,\ldots,\,N-1$}
    \If{$n_{\mathrm{min}} \leq N - i \leq n_{\mathrm{max}}$}
        \Comment{Editing window}
        \State $\xsrci \leftarrow (1-t_i)\,\xsrc + t_i\,\bm{\epsilon}_{i}$, \quad $\bm{\epsilon}_{i} \sim \mathcal{N}(\bm{0},\bm{I})$
        \State $\xtari \leftarrow \bm{x}_{t_i} + \xsrci - \xsrc$
        \State
            $\Dv_{t_i} \leftarrow 
            \vtheta(\xtari,\,t_i;\,\etar) -
            \vtheta(\xsrci,\,t_i;\,\esrc) $
            
        \State $\vsem \leftarrow \vtheta(\xsrci,\,t_i;\,\etar) - \vtheta(\xsrci,\,t_i;\,\esrc)$
            
        \State $\vlow \leftarrow \LF(\vsem)$
            \Comment{$L$-level Haar extraction, as in Eq.~\eqref{eq:lf_op}}
        \State $ {\widetilde{\Dv}}_{t_i}\leftarrow {{\Dv}}_{t_i} + \lambda \cdot t_i^2 \cdot \vlow$
        \State ${\widetilde{\Dv}}_{t_i} \leftarrow M \odot {\widetilde{\Dv}}_{t_i}$
        \State $\bm{x}_{t_{i+1}} \leftarrow \bm{x}_{t_i} + (t_{i+1} - t_i)\,{\widetilde{\Dv}}_{t_i}$
    \ElsIf{$N - i < n_{\mathrm{min}}$}
        \Comment{Standard denoising}
        \State $\bm{x}_{t_{i+1}} \leftarrow \bm{x}_{t_i}
               + (t_{i+1}-t_i)\,\vtheta(\bm{x}_{t_i},\,t_i;\,\etar)$

    \EndIf
    
\EndFor
\State \Return $\bm{x}_{t_N}$
\end{algorithmic}
\end{algorithm}
\section{Experiments}
In this section, we evaluate the performance of our proposed method against inversion-based and inversion-free baselines. We conduct comprehensive experiments to demonstrate the effectiveness of our approach in terms of editing quality, structure preservation, and background consistency.
\subsection{Setup}

Our method is implemented on FLUX~\cite{flux2024}, Stable Diffusion 3 (SD3)~\cite{esser2024scaling}, and SD3.5. We report the main quantitative and qualitative results on FLUX, with additional SD3 and SD3.5 results in Appendices~B and~C.
\paragraph{Dataset and Baselines} For evaluation, we utilize PIE-Bench~\cite{jupnp}, consisting of 700 source images and target prompts organized into ten diverse editing scenarios. Additional high-resolution evaluation on EditBench~\cite{lin2024schedule} is provided in Appendix~\ref{sec:qual_flux}. We compare our method against a comprehensive set of baselines. For diffusion-based methods, we compare with 
PnP~\cite{tumanyan2023plug} and 
MasaCtrl~\cite{cao2023masactrl}. Additionally, we evaluate FreeDiff~\cite{wu2024freediff}, which employs progressive frequency truncation to preserve structural consistency by utilizing low-frequency components from the source image during the editing process. For the inversion-based method, we compare with RF-Inv~\cite{rout2025semantic}, StableFlow~\cite{avrahami2025stable}, RF-Edit~\cite{wang2025taming}, FireFlow~\cite{deng2025fireflow} and DNA-Edit~\cite{xie2025dnaedit}. For inversion-free methods, we compare with FlowEdit~\cite{kulikov2024flowedit} in the main FLUX evaluation, and provide additional comparisons with FlowAlign~\cite{kim2025flowalign} and DVRF~\cite{beaudouin2025delta} in Appendix~\ref{sec:exp_sd3}. Comparisons with recent frequency-aware methods, including FSI-Edit~\cite{teng2024fsi} and FIA-Edit~\cite{liu2024fia}, are reported in Appendix~\ref{sec:exp_sd35}. Runtime analysis is provided in Appendix~\ref{sec:runtime}.

\paragraph{Metrics}
We report three categories of metrics: Structural consistency, background preservation, and text-image consistency. Structure distance is abbreviated as Struct. in~\cref{tab:pie}, measuring structural similarity via the cosine similarity between DINO~\cite{caron2021emerging} feature self-similarity matrices. Background preservation evaluates non-edited regions using peak signal-to-noise ratio (PSNR)~\cite{huynh2008scope},
structural similarity index measure (SSIM)~\cite{wang2004image},
mean squared error (MSE),
and learned perceptual image patch similarity (LPIPS)~\cite{zhang2018unreasonable}, computed outside the annotated editing masks. For text-image consistency, we compute CLIP~\cite{radford2021learning} similarity on both the full image and the edited region to assess global alignment and local editing accuracy.

\paragraph{Implementation Details}

We follow the FlowEdit~\cite{kulikov2024flowedit} setup for the sampling configuration.
Specifically, we adopt $N = 28$ sampling steps with the editing window defined as
$n_{\mathrm{min}} = 0$ and $n_{\mathrm{max}} = 24$ for FLUX. We set the semantic strength parameter $\lambda = 3.0$ and the wavelet level to $L = 3$ for FLUX.

\subsection{Quantitative Results}
As shown in~\cref{tab:pie}, our method achieves the best CLIP scores while also leading across background preservation metrics. Although DNA-Edit obtains a strong structure distance, its lower CLIP scores indicate weaker semantic editing. In contrast, our approach provides stronger target alignment while maintaining higher PSNR, lower LPIPS/MSE, and higher SSIM, showing that wavelet-guided compensation better balances edit strength and background fidelity in inversion-free editing.

\begin{table}[t]
\caption{
Quantitative comparison on the PIE-Bench. 
Comparison across all methods without grouping by backbone.
The best and second-best results are shown in \textbf{bold}
and \underline{underline}, respectively.
}
\label{tab:pie}
\centering
\setlength{\tabcolsep}{4.6pt}
\resizebox{0.95\textwidth}{!}{%
\begin{tabular}{llccccccc}

% \begin{tabular}{l@{\hspace{0.5em}}l@{\hspace{0.5em}}c@{\hspace{0.5em}}cccc@{\hspace{0.5em}}cc}
\toprule
\multirow{2}{*}{Method} & \multirow{2}{*}{Model} & \multirow{2}{*}{\makecell{Struct.$\times 10^3$ \\ $\downarrow$}} 
& \multicolumn{4}{c}{\makecell{Background \\ Preservation}} 
& \multicolumn{2}{c}{\makecell{CLIP \\ Similarity}} \\
\cmidrule(lr){4-7} \cmidrule(lr){8-9}
& & 
& PSNR $\uparrow$ 
& LPIPS$\times 10^3$ $\downarrow$ 
& MSE$\times 10^4$ $\downarrow$ 
& SSIM$\times 10^2$ $\uparrow$ 
& Whole $\uparrow$ 
& Edited $\uparrow$ \\
\midrule

PnP~\cite{tumanyan2023plug}& Diffusion & 23.47 & 22.48 & 105.70 & 79.83 & 80.22 & \underline{25.44} & 22.60 \\
MasaCtrl~\cite{cao2023masactrl} & Diffusion & 23.68 & 22.66 & 87.54 & 80.66 & 81.89 & 24.37 & 21.36 \\
FreeDiff~\cite{wu2024freediff} & Diffusion & \underline{17.98} & 24.78 & 89.02 & 54.81 & 81.92 & 25.16 & 22.15 \\
\midrule
RF-Inv~\cite{rout2025semantic} & FLUX & 64.61 & 17.97 & 242.00 & 221.24 & 64.92 & 25.29 & 22.91 \\
StableFlow~\cite{avrahami2025stable} & FLUX & 19.25 & 23.04 & \underline{77.09} & 84.78 & 87.22 & 24.06 & 21.18 \\
RF-Edit~\cite{wang2025taming} & FLUX & 24.45 & 24.41 & 113.44 & 56.46 & 83.84 & 25.03 & 22.28 \\
FireFlow~\cite{deng2025fireflow} & FLUX & 27.27 & 23.08 & 128.65 & 71.11 & 81.25 & 25.34 & \underline{22.92} \\

FlowEdit~\cite{kulikov2024flowedit} & FLUX & 27.61 & 22.23 & 111.32 & 90.98 & 83.64 & 25.39 & 22.62 \\
DNA-Edit~\cite{xie2025dnaedit} & FLUX & \textbf{16.83} & \underline{25.21} & 86.48 & \underline{48.11} & \underline{87.23} & 24.82 & 22.12 \\
Ours & FLUX & 30.95 & \textbf{26.17} & \textbf{54.02} & \textbf{39.24} & \textbf{90.60} & \textbf{25.52} & \textbf{23.05} \\

\bottomrule
\end{tabular}

}
\end{table}

\subsection{Qualitative Results}

\begin{figure}[t]
    \centering
    \includegraphics[width=0.90\linewidth]{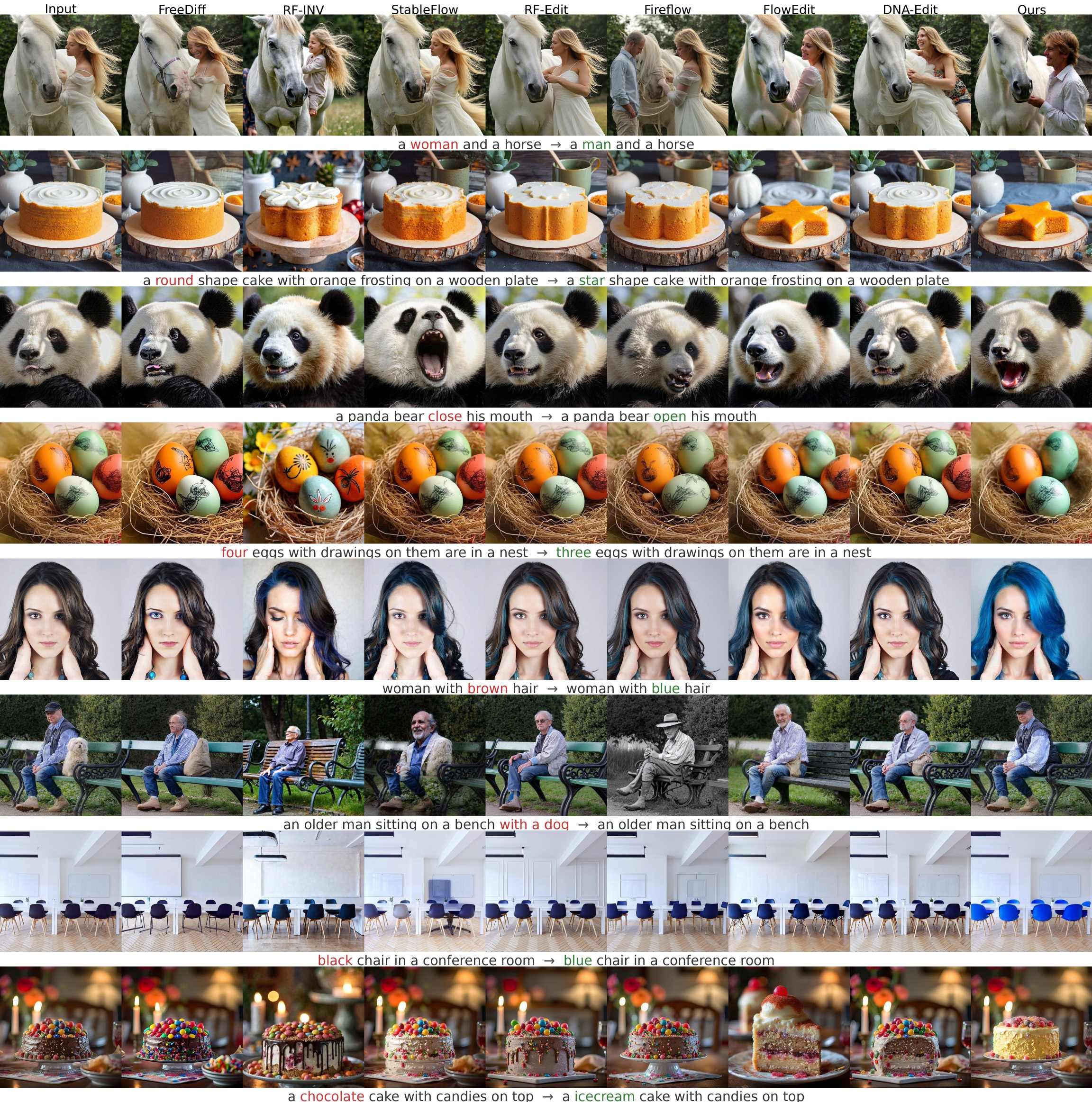}
    \caption{Qualitative comparison on the PIE-Bench.}
    \label{fig:quali}
\end{figure}

\cref{fig:quali} presents visual comparisons across a range of editing tasks, including both local attribute manipulation and more complex structural modifications. As illustrated in the figure, our method achieves precise semantic modification while preserving the identity and overall structure of the original image. For example, in the open-mouth panda case, the mouth region is successfully modified to reflect the target instruction, while the identity, pose, and surrounding environment of panda remain largely unchanged. Similar behavior can be observed in other examples, where the target attributes are effectively realized while preserving the main content
of the original image. These results demonstrate that our approach effectively balances semantic editing strength and structural consistency across diverse editing scenarios.

\subsection{Sensitivity and Design Choices}
Our method introduces two design parameters, namely the wavelet level \(L\) and semantic strength \(\lambda\). As summarized in~\cref{tab:hparam_sensitivity}, increasing \(L\)  generally favors structural fidelity and background preservation, while \(\lambda\) controls the preservation-editability trade-off. We use \(L=3\) and  \(\lambda=3.0\) as the default FLUX setting. \cref{tab:band_selection} shows that low-frequency compensation performs better than full-spectrum or high-frequency-only injection. Complete grids for \(\lambda\) and \(L\) are provided in Appendix~\ref{sec:FLUX_parameter}, with low-pass operator and schedule ablations in Appendices~\ref{sec:freq_ablation} and~\ref{sec:time_ablation}.

\begin{table}[!t]
\caption{
Ablation study of the wavelet decomposition level \(L\) and the semantic
strength parameter \(\lambda\) on PIE-Bench under FLUX.
We report one-factor sensitivity slices for \(L\) and \(\lambda\).
Our final choice (\(L=3\), \(\lambda=3.0\)) is highlighted in gray.
The complete hyperparameter grid is provided in Appendix.
}
\label{tab:hparam_sensitivity}
\centering
\setlength{\tabcolsep}{4pt}
\renewcommand{\arraystretch}{0.96}
\resizebox{0.95\linewidth}{!}{
\begin{tabular}{lcc c cccc cc}
\toprule
\multirow{2}{*}{Setting}
& \multirow{2}{*}{\(L\)}
& \multirow{2}{*}{\(\lambda\)}
& \multirow{2}{*}{\makecell{Struct.\(\times 10^3\)\\\(\downarrow\)}}
& \multicolumn{4}{c}{\makecell{Background\\Preservation}}
& \multicolumn{2}{c}{\makecell{CLIP\\Similarity}} \\
\cmidrule(lr){5-8}\cmidrule(lr){9-10}
& & &
& PSNR \(\uparrow\)
& LPIPS\(\times10^3\) \(\downarrow\)
& MSE\(\times10^4\) \(\downarrow\)
& SSIM\(\times10^2\) \(\uparrow\)
& Whole \(\uparrow\)
& Edited \(\uparrow\) \\
\midrule
\(L\) sweep (\(\lambda=3.0\))
& 1 & 3.0 & 35.31 & 25.26 & 58.48 & 46.12 & 89.95 & 25.56 & 23.01 \\
& 2 & 3.0 & 34.21 & 25.51 & 57.36 & 44.14 & 90.16 & 25.59 & 23.01 \\
\rowcolor{gray!15}
& 3 & 3.0 & 30.95 & 26.17 & 54.02 & 39.24 & 90.60 & 25.52 & 23.05 \\
& 4 & 3.0 & 26.06 & 27.05 & 49.12 & 33.36 & 91.17 & 25.39 & 22.90 \\
\midrule
\(\lambda\) sweep (\(L=3\))
& 3 & 2.0 & 25.72 & 26.88 & 49.64 & 34.09 & 91.16 & 25.48 & 22.98 \\
& 3 & 2.5 & 28.30 & 26.52 & 51.83 & 36.59 & 90.88 & 25.54 & 23.05 \\
\rowcolor{gray!15}
& 3 & 3.0 & 30.95 & 26.17 & 54.02 & 39.24 & 90.60 & 25.52 & 23.05 \\
& 3 & 3.5 & 33.67 & 25.80 & 56.24 & 41.99 & 90.33 & 25.55 & 23.09 \\
& 3 & 4.0 & 36.65 & 25.44 & 58.68 & 45.00 & 90.04 & 25.44 & 22.94 \\
\bottomrule
\end{tabular}
}
\end{table}

\begin{table}[!t]
\caption{Frequency-band selection on PIE-Bench under FLUX.}
\label{tab:band_selection}
\centering
\setlength{\tabcolsep}{4pt}
\renewcommand{\arraystretch}{0.96}
\resizebox{0.95\linewidth}{!}{
\begin{tabular}{l c cccc cc}
\toprule
\multirow{2}{*}{Method}
& \multirow{2}{*}{\makecell{Struct.\(\times 10^3\)\\\(\downarrow\)}}
& \multicolumn{4}{c}{\makecell{Background\\Preservation}}
& \multicolumn{2}{c}{\makecell{CLIP\\Similarity}} \\
\cmidrule(lr){3-6}\cmidrule(lr){7-8}
&
& PSNR \(\uparrow\)
& LPIPS\(\times10^3\) \(\downarrow\)
& MSE\(\times10^4\) \(\downarrow\)
& SSIM\(\times10^2\) \(\uparrow\)
& Whole \(\uparrow\)
& Edited \(\uparrow\) \\
\midrule
Zero (no comp.)       & 27.61 & 22.23 & 111.32 & 90.98 & 83.64 & 25.39 & 22.62 \\
Identity (full)       & 41.53 & 23.61 & 69.77  & 63.31 & 88.05 & 25.03 & 22.24 \\
Highpass only         & 33.20 & 24.81 & 59.69  & 50.37 & 89.25 & 24.97 & 22.34 \\
\textbf{Haar wavelet} & 30.95 & 26.17 & 54.02  & 39.24 & 90.60 & 25.52 & 23.05 \\
\bottomrule
\end{tabular}
}
\end{table}

\begin{table}[!t]
\centering
\small
\setlength{\tabcolsep}{3pt}
\caption{User preference rate (\%).}
\label{tab:user_study}
\begin{tabular}{lcc|lcc}
\toprule
Method & Ours & Base & Method & Ours & Base \\
\midrule
FireFlow     & 71 & 29 & FreeDiff      & 58 & 42 \\
StableFlow   & 68 & 32 & DNA-Edit      & 58 & 42 \\
FlowEdit     & 66 & 34 & RF-Edit     & 58 & 42 \\
RF-Inv & 64 & 36 \\
\bottomrule
\end{tabular}
\end{table}

\subsection{User Study}

We further assess perceptual quality through a pairwise user study. For each baseline comparison, we randomly select 100 edited image pairs generated by our method and one of seven baselines using the same source image and edit prompt. Ten participants choose the result that better preserves unedited regions while reflecting the target edit. As shown in~\cref{tab:user_study}, our method is preferred over all baselines, complementing the quantitative results.

\section{Conclusion}
In this paper, we investigate inversion-free image editing from the perspective of trajectory dynamics in rectified flow models. Through empirical and analytical observations, we identified that under certain global attribute shifts, the semantic editing signal may become weak in the early timesteps due to strong geometric coupling along the source trajectory. Motivated by this analysis, we introduced a time-dependent compensation mechanism that enhances semantic deviation when the trajectory is insufficiently responsive to the target prompt, while remaining compatible with the original formulation. The proposed strategy improves editing strength while maintaining background fidelity. Extensive experiments demonstrate that our method achieves more faithful semantic modifications and stronger perceptual preference across diverse editing scenarios.

\clearpage

\appendices
\begin{center}
  {\Large\bfseries
   Wavelet-Guided Semantic Signal Compensation\\
   for Inversion-Free Image Editing\\[0.3em]
   Appendices\par}
\end{center}

The appendices are organized as follows. Appendix~\ref{sec:prem_appen} provides additional preliminaries on FlowEdit and the two-dimensional Haar wavelet transform. Appendix~\ref{sec:exp_sd3} reports additional SD3 experiments, including quantitative results, qualitative comparisons, and hyperparameter ablations. Appendix~\ref{sec:exp_sd35} further presents SD3.5 results and comparisons with recent frequency-aware editing methods. Appendix~\ref{sec:qual_flux} provides additional FLUX qualitative results and high-resolution evaluation on EditBench. Appendix~\ref{sec:FLUX_parameter} contains detailed FLUX ablation studies on hyperparameters, frequency-band selection, low-pass operator choices, and time-dependent weighting schedules. Appendix~\ref{sec:runtime} reports the inference-time comparison. Finally, Appendix~\ref{sec:user} provides the details and complete statistics of the user study.

\section{Preliminaries on FlowEdit}
\label{sec:prem_appen}
This section provides additional details to clarify the formulations used in the main text. We describe the translation-coupled trajectory construction of FlowEdit and the Haar wavelet transform with more explicit notation.

\subsection{FlowEdit}
\label{sec:flowedit_appendix}

This section presents FlowEdit~\cite{kulikov2024flowedit} from the perspective of the edited-path update,
using exactly the same notation as Algorithm 1. Throughout, $\boldsymbol{x}_{t_i}$ does not represent the noisy latent at timestep \(t_i\); rather, it represents the evolving noise-free edited latent state (analogous to the direct path $\bm{Z}_t^{\mathrm{FE}}$ in the FlowEdit~\cite{kulikov2024flowedit}). Thus, it is initialized as $\boldsymbol{x}_{t_0} = \boldsymbol{x}^{\mathrm{src}}$. Given a source latent $\bx^{\mathrm{src}}$, at each timestep $t_i$ we sample Gaussian noise
$\bm{\epsilon}_{i} \sim \mathcal{N}(\bm{0},\bm{I})$ and construct the source-side noisy query
\begin{equation}
\bx^{\mathrm{src}}_{t_i}
=
(1-t_i)\,\bx^{\mathrm{src}} + t_i\,\bm{\epsilon}_{i}.
\label{eq:flowedit_src_traj_appendix}
\end{equation}

Using the same noise realization, FlowEdit forms the translation-coupled target query point
\begin{equation}
{\bx}^{\mathrm{tar}}_{t_i}
=
{\bx}_{t_i} + \bx^{\mathrm{src}}_{t_i} - \bx^{\mathrm{src}}.
\label{eq:flowedit_translated_query_appendix}
\end{equation}
Let $\esrc$ and $\etar$ be the text embeddings of the source and target prompts.
The geometric editing signal (delta velocity) is evaluated as
\begin{equation}
\Delta \bv_{t_i}
\triangleq
\bv_{\btheta}\!\left({\bx}^{\mathrm{tar}}_{t_i},\, t_i;\, \etar\right)
-
\bv_{\btheta}\!\left({\bx}^{\mathrm{src}}_{t_i},\, t_i;\, \esrc\right),
\label{eq:flowedit_delta_v_appendix}
\end{equation}
where we keep the notation $\bv_{\btheta}(\cdot, t;\, \be)$ to denote the rectified-flow velocity field
parameterized by $\btheta$ and conditioned on the text embedding $\be$.

Optionally, one may approximate the noise-averaged update direction at each timestep $t_i$ by sampling
multiple noises $\{\bm{\epsilon}_{i,k}\}_{k=1}^{K}$ within the same timestep and averaging the resulting
delta velocities. Concretely, for each fixed $i$, we draw $\bm{\epsilon}_{i,k} \sim \mathcal{N}(\bm{0},\bm{I})$
for $k=1,\ldots,K$, and evaluate $\Delta \bv_{t_i}(\bm{\epsilon}_{i,k})$ using the same construction as
Eqs.~\eqref{eq:flowedit_src_traj_appendix}--\eqref{eq:flowedit_delta_v_appendix} (i.e., the source-side and target-side
network queries share the same $\bm{\epsilon}_{i,k}$). For a discrete timestep schedule $\{t_i\}_{i=0}^{N}$, the edited latent is updated via Euler discretization:
\begin{equation}
\bx_{t_{i+1}}
=
\bx_{t_i}
+
(t_{i+1}-t_i)\,
\frac{1}{K}
\sum_{k=1}^{K}
\Delta \bv_{t_i}\!\left(\bm{\epsilon}_{i,k}\right),
\qquad i=0,1,\ldots,N-1.
\label{eq:flowedit_update_appendix}
\end{equation}
In our Algorithm 1, we follow the standard FlowEdit setting and use a single noise sample per timestep, i.e., $K=1$. The main paper also presents FlowEdit using a coupled dynamics viewpoint,
\(\mathrm{d}\bx^{\mathrm{tar}}_t = \mathrm{d}\bx^{\mathrm{src}}_t + \Delta \bv_t\,\mathrm{d}t\),
whose discretization highlights a source transport term. This expression is an interpretive decomposition explaining why the method tends to preserve source structure under
translation coupling. In our implementation, we do not separately
integrate a source ODE; instead, structure preservation is realized implicitly through the translated query
\(\bx_{t_i} + \bx^{\mathrm{src}}_{t_i} - \bx^{\mathrm{src}}\) and the shared noise realization in
Eqs.~\eqref{eq:flowedit_translated_query_appendix}--\eqref{eq:flowedit_update_appendix}.
Although we do not explicitly integrate a separate source ODE, the translated target query defined in Eq.~\eqref{eq:flowedit_translated_query_appendix} satisfies the following algebraically equivalent recursion:
\begin{equation}
\bx^{\mathrm{tar}}_{t_{i+1}}
=
\bx^{\mathrm{tar}}_{t_i}
+
\left(\bx^{\mathrm{src}}_{t_{i+1}}-\bx^{\mathrm{src}}_{t_i}\right)
+
(t_{i+1}-t_i)\,
\frac{1}{K}
\sum_{k=1}^{K}
\Delta \bv_{t_i}\!\left(\bm{\epsilon}_{i,k}\right).
\label{eq:flowedit_transport_equiv_appendix}
\end{equation}
That is, the source-transport viewpoint follows directly from the translated-query construction and the edited-path Euler update, rather than from a separately integrated source ODE.

\subsection{Two-Dimensional Haar Wavelet Transform}
\label{sec:haar_prem}

Let $X \in \RR^{H \times W}$ denote a single-channel spatial tensor. The transform is applied
identically and independently to each channel of the multi-channel latent. For every
non-overlapping $2 \times 2$ block indexed by
$i \in \{0,\ldots,H/2-1\}$ and $j \in \{0,\ldots,W/2-1\}$, denote the four co-located
samples as $a = X[2i,2j]$, $b = X[2i,2j+1]$, $c = X[2i+1,2j]$, and $d = X[2i+1,2j+1]$.
The single-level forward Haar transform produces four sub-bands of spatial size
$(H/2) \times (W/2)$:
\begin{align}
    cA[i,j] &= \tfrac{1}{2}(a+b+c+d), \label{eq:cA}\\
    cH[i,j] &= \tfrac{1}{2}(a-b+c-d), \label{eq:cH}\\
    cV[i,j] &= \tfrac{1}{2}(a+b-c-d), \label{eq:cV}\\
    cD[i,j] &= \tfrac{1}{2}(a-b-c+d). \label{eq:cD}
\end{align}
Here $cA$ encodes the local spatial average, while $cH$, $cV$, and $cD$ encode horizontal,
vertical, and diagonal local differences, respectively. The single-level inverse transform
reconstructs the original four samples:
\begin{align}
    X[2i,\;2j]     &= \tfrac{1}{2}(cA+cH+cV+cD), \label{eq:idwt_a}\\
    X[2i,\;2j+1]   &= \tfrac{1}{2}(cA-cH+cV-cD), \label{eq:idwt_b}\\
    X[2i+1,\;2j]   &= \tfrac{1}{2}(cA+cH-cV-cD), \label{eq:idwt_c}\\
    X[2i+1,\;2j+1] &= \tfrac{1}{2}(cA-cH-cV+cD). \label{eq:idwt_d}
\end{align}
Substituting Eqs.~\eqref{eq:cA}--\eqref{eq:cD} into Eqs.~\eqref{eq:idwt_a}--\eqref{eq:idwt_d}
recovers $a$, $b$, $c$, and $d$ exactly. The forward and inverse transforms therefore allow exact reconstruction.
By direct substitution, $X[2i,2j]
    % &= \frac{1}{2}
    %   \cdot
    %   \frac{(a+b+c+d)+(a-b+c-d)+(a+b-c-d)+(a-b-c+d)}{2}
    = a.$
Analogous computations yield $X[2i,2j+1] = b$, $X[2i+1,2j] = c$, and
$X[2i+1,2j+1] = d$.

\paragraph{Practical Considerations}
When $H$ or $W$ is odd, a single-pixel reflect-padding is applied before each forward transform
step, and the corresponding border row or column is cropped after reconstruction. The number
of decomposition levels is bounded by $L \leq \lfloor \log_2 \min(H,W) \rfloor$ to prevent
any sub-band from collapsing to a single spatial location. All operations are implemented via
standard tensor indexing, are fully differentiable, and execute natively on GPU hardware.

\section{Experimental Results for SD3}
\label{sec:exp_sd3}
We compare with FlowEdit~\cite{kulikov2024flowedit}, FlowAlign~\cite{kim2025flowalign},
and DVRF~\cite{beaudouin2025delta}.
We follow the FlowEdit~\cite{kulikov2024flowedit} setup for the sampling configuration on PIE-Bench.
Specifically, we adopt $N = 50$ sampling steps with the editing window defined as
$n_{\mathrm{min}} = 0$ and $n_{\mathrm{max}} = 33$ for SD3. We use $N=50$ Euler updates for SD3, with timesteps indexed by $i=0,\dots,49$ and a scheduler descending from $t_0\approx 1$ to $t_{50}\approx 0$. Following FlowEdit, the editing window is specified by the reverse step index $s_i=N-i$. With $n_{\mathrm{min}} = 0$ and $n_{\mathrm{max}} = 33$, the highest-noise prefix $i=0,\dots,16$ is skipped via identity updates $\bm{x}_{t_{i+1}} \leftarrow \bm{x}_{t_i}$, and the editing update is applied for $i=17,\dots,49$. Hence, no final standard-sampling stage is used in this setting. Our method introduces two additional hyperparameters: the semantic strength parameter $\lambda = 2.2$ and the wavelet decomposition level $L=4$.

\subsection{Quantitative Results}
As shown in \cref{tab:sd3_quantitative}, we present the quantitative comparison on PIE-Bench with SD3.
Our method achieves the best results on PSNR, LPIPS, MSE, SSIM, and Edited CLIP similarity. These results indicate that our method is particularly effective at preserving background appearance during editing, without sacrificing semantic consistency in the edited region. 
In particular, compared with FlowAlign, our method consistently improves all background-preservation metrics by a clear margin, demonstrating the advantage of our frequency-aware design in maintaining low-frequency content and suppressing undesired distortions. Compared with DVRF, our method yields slightly weaker Struct. and Whole CLIP similarity, but provides substantially better reconstruction fidelity and stronger semantic alignment in edited regions.
This suggests that our method strikes a more favorable trade-off between preservation and editing quality. Overall, the quantitative results verify that our method achieves more balanced performance among all compared approaches.

\begin{table}[!t]
\caption{
Quantitative comparison on the PIE-Bench with SD3. 
Comparison across all methods without grouping by backbone.
The best and second-best results are shown in \textbf{bold}
and \underline{underline}, respectively. Struct. denotes structure distance.
}
\label{tab:sd3_quantitative}
\centering
\setlength{\tabcolsep}{5pt}
\resizebox{0.9\linewidth}{!}{
\begin{tabular}{lccccccc}
\toprule
\multirow{2}{*}{Method} & 
\multirow{2}{*}{\makecell{Struct.$\times 10^3$ \\ $\downarrow$}} 
& \multicolumn{4}{c}{\makecell{Background \\ Preservation}} 
& \multicolumn{2}{c}{\makecell{CLIP \\ Similarity}} \\
\cmidrule(lr){3-6} \cmidrule(lr){7-8}
& 
& PSNR $\uparrow$ 
& LPIPS$\times 10^3$ $\downarrow$ 
& MSE$\times 10^4$ $\downarrow$ 
& SSIM$\times 10^2$ $\uparrow$ 
& Whole $\uparrow$ 
& Edited $\uparrow$ \\
\midrule

FlowEdit~\cite{kulikov2024flowedit} & 27.43 & 22.20 & 105.04 & 86.93 & 83.52 & \underline{26.61} & 23.68 \\
DVRF~\cite{beaudouin2025delta} & \textbf{23.21} & 23.37 & 93.52 & 67.25 & 84.82 & \textbf{26.62} & \underline{23.77} \\
FlowAlign~\cite{kim2025flowalign} & 33.43 & \underline{23.94} & \underline{72.76} & \underline{56.70} & \underline{86.11} & 25.63 & 22.35 \\
Ours  & \underline{26.20} & \textbf{25.26} & \textbf{52.46} & \textbf{41.60} & \textbf{89.80} & 26.44 & \textbf{23.97} \\

\bottomrule
\end{tabular}
}
\end{table}

\subsection{Qualitative Comparisons}

\begin{figure*}[!t]
    \centering
    \includegraphics[width=0.95\textwidth]{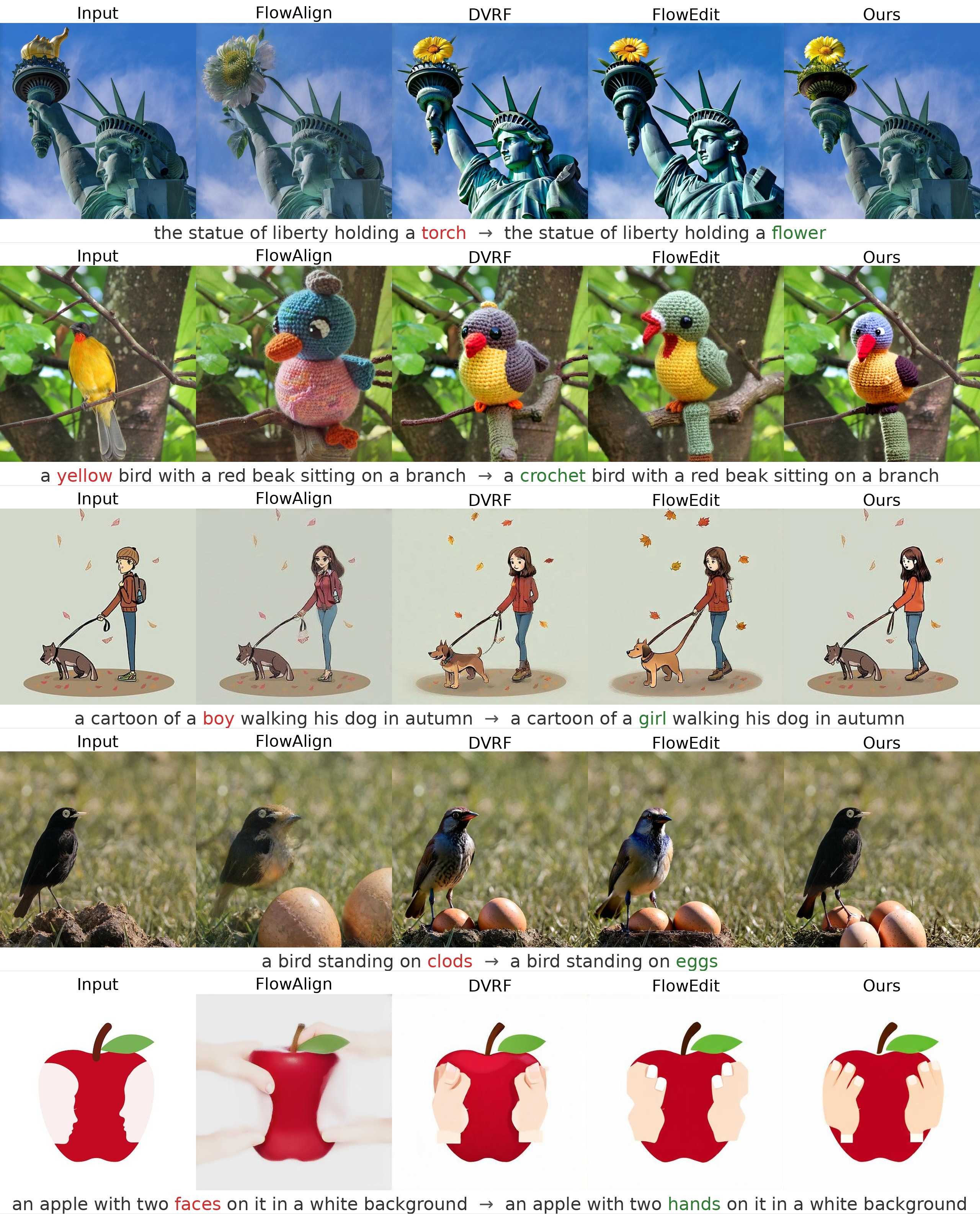}
    \caption{Qualitative comparison on PIE-Bench with SD3 (Part 1).}
    \label{fig:qualitative_results_part1}
\end{figure*}

\begin{figure*}[!t]
    \centering
    \includegraphics[width=0.95\textwidth]{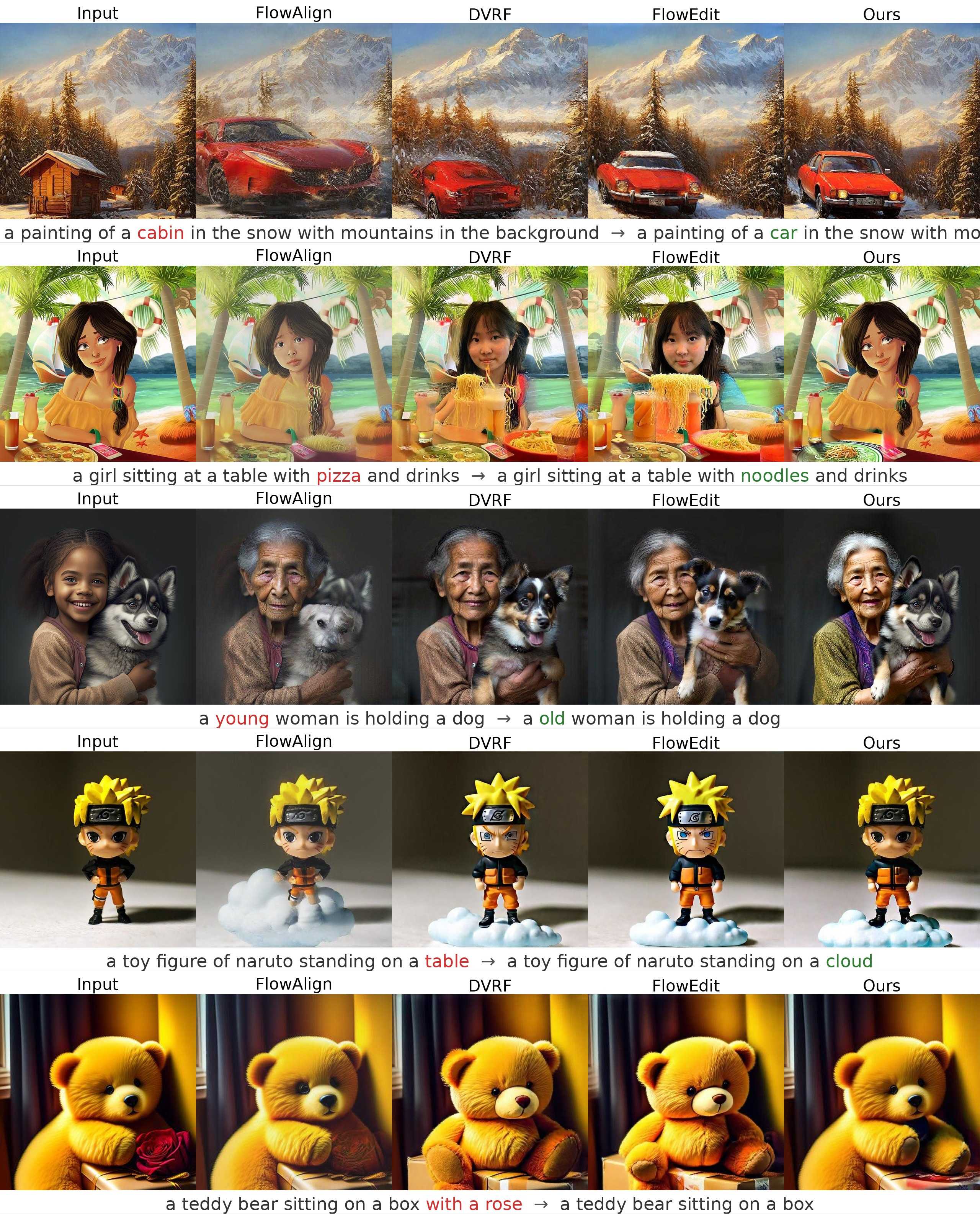}
    \caption{Qualitative comparison on PIE-Bench with SD3 (Part 2).}
    \label{fig:qualitative_results_part2}
\end{figure*}

As shown in~\cref{fig:qualitative_results_part1,fig:qualitative_results_part2}, our method consistently achieves more accurate semantic editing while preserving the visual structure and identity cues of the input image. Compared with FlowEdit, FlowAlign, and DVRF, the proposed method produces edits that more faithfully match the target prompt across a wide range of semantic transformations, including object replacement, attribute modification, human character substitution, and background-aware content editing. 

In the first two examples, the method successfully replaces the torch with a flower and transforms a yellow bird into a crochet bird, while maintaining the global composition and the local spatial arrangement of the original scene. In the cartoon and portrait examples, the method also performs reliable category-level and age-related edits, changing a boy into a girl and a young woman into an old woman without introducing severe geometric distortion or identity inconsistency. Similar advantages can be observed in the examples involving clods to eggs, faces to hands, cabin to car, pizza to noodles, table to cloud, and rose removal. 

A notable property of the proposed method is the balance between editability and content preservation. Existing baselines often suffer from either under-editing, where the target concept is only partially realized, or over-editing, where irrelevant regions are unnecessarily modified. In contrast, our results show that the edited content is well aligned with the target semantics, while unedited regions such as pose, viewpoint, illumination, and scene layout remain largely stable. These observations suggest that the proposed semantic strength control and wavelet-based design improve both the precision of semantic manipulation and the preservation of fine visual details.

\subsection{Ablation Study with SD3}

Our method introduces two additional hyperparameters: the semantic strength parameter $\lambda$ and the wavelet decomposition level $L$.
To study their effects, we conduct an ablation analysis on PIE-Bench under SD3 by varying $L \in \{1,2,3,4\}$ and $\lambda \in \{1.2, 1.4, 1.6, 1.8, 2.0, 2.2, 2.4, 2.5, 3.0, 3.5, 4.0\}$.

As shown in~\cref{tab:ablation_LB_SD3}, increasing $L$ consistently improves structural fidelity and background preservation, suggesting that higher wavelet decomposition levels are more effective at preserving low-frequency content.
In contrast, increasing $\lambda$ generally enhances semantic editing strength but gradually compromises structural consistency and background fidelity.
Moreover, the gain in semantic alignment becomes limited when $\lambda$ is overly large, indicating a saturation effect.

Overall, the results show a clear trade-off between preservation quality and semantic editing performance.
Therefore, we adopt $(L,\lambda)=(4,2.2)$ as the default setting, since it provides the best balance between faithful preservation and effective editing.

\begin{figure}[t]
    \centering
    \includegraphics[width=0.95\linewidth]{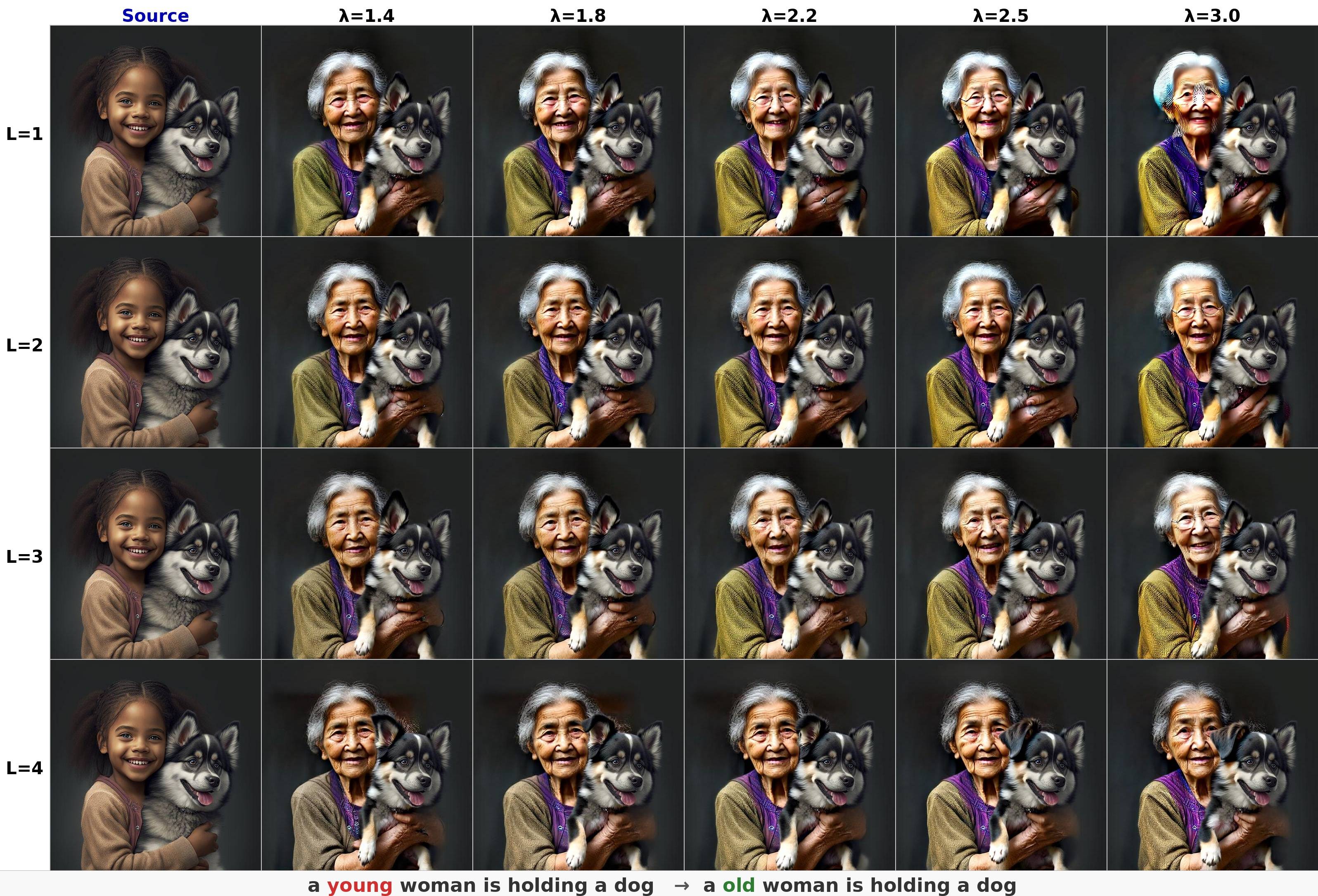}
    \caption{Qualitative ablation on $L$ and $\lambda$ under SD3.}
    \label{fig:ablation_SD3_vis}
\end{figure}

The qualitative examples in~\cref{fig:ablation_SD3_vis} show a similar trend, with nearby configurations yielding visually comparable results and moderate $\lambda$ providing a favorable balance between edit strength and visual fidelity.

\begin{table}[t]
\caption{
Ablation study of the wavelet decomposition level $L$ and the semantic strength parameter $\lambda$ on PIE-Bench under SD3.
The best and second-best results are shown in \textbf{bold} and \underline{underline}, respectively.
Our final choice is highlighted in gray.
}
\label{tab:ablation_LB_SD3}
\centering
\setlength{\tabcolsep}{6pt}
\resizebox{0.80\linewidth}{!}{
\begin{tabular}{cc c cccc cc}
\toprule
\multirow{2}{*}{$L$} & \multirow{2}{*}{$\lambda$} & \multirow{2}{*}{\makecell{Struct.$\times 10^3$ \\ $\downarrow$}}
& \multicolumn{4}{c}{\makecell{Background \\ Preservation}}
& \multicolumn{2}{c}{\makecell{CLIP \\ Similarity}} \\
\cmidrule(lr){4-7} \cmidrule(lr){8-9}
& &
& PSNR $\uparrow$
& LPIPS$\times 10^3$ $\downarrow$
& MSE$\times 10^4$ $\downarrow$
& SSIM$\times 10^2$ $\uparrow$
& Whole $\uparrow$
& Edited $\uparrow$ \\
\midrule
1 & 1.2 & 28.41 & 24.77 & 54.99 & 45.60 & 89.55 & 26.46 & 23.92 \\
1 & 1.4 & 29.43 & 24.59 & 56.08 & 47.27 & 89.44 & 26.46 & 24.00 \\
1 & 1.6 & 30.58 & 24.40 & 57.17 & 49.01 & 89.32 & 26.46 & 23.96 \\
1 & 1.8 & 31.79 & 24.22 & 58.42 & 51.04 & 89.19 & 26.48 & 23.95 \\
1 & 2.0 & 32.98 & 24.01 & 59.70 & 53.18 & 89.06 & 26.51 & 23.98 \\
1 & 2.2 & 34.31 & 23.82 & 61.00 & 55.42 & 88.92 & 26.52 & 23.89 \\
1 & 2.4 & 35.75 & 23.63 & 62.44 & 57.91 & 88.78 & 26.52 & 23.83 \\
1 & 2.5 & 36.50 & 23.51 & 63.25 & 59.23 & 88.69 & \underline{26.54} & 23.89 \\
1 & 3.0 & 40.27 & 23.02 & 66.90 & 66.00 & 88.31 & 26.50 & 23.78 \\
1 & 3.5 & 44.25 & 22.55 & 70.48 & 72.91 & 87.94 & 26.38 & 23.61 \\
1 & 4.0 & 48.36 & 22.09 & 74.42 & 80.43 & 87.55 & 26.28 & 23.44 \\
\midrule

2 & 1.2 & 27.36 & 24.97 & 53.96 & 44.04 & 89.67 & 26.43 & 23.99 \\
2 & 1.4 & 28.28 & 24.82 & 54.88 & 45.32 & 89.58 & 26.46 & 23.95 \\
2 & 1.6 & 29.28 & 24.66 & 55.87 & 46.75 & 89.48 & 26.48 & 23.98 \\
2 & 1.8 & 30.28 & 24.48 & 56.93 & 48.37 & 89.37 & 26.45 & 23.98 \\
2 & 2.0 & 31.34 & 24.32 & 58.06 & 50.13 & 89.26 & \textbf{26.55} & 23.95 \\
2 & 2.2 & 32.53 & 24.14 & 59.23 & 52.00 & 89.14 & 26.49 & 23.96 \\
2 & 2.4 & 33.70 & 23.97 & 60.42 & 53.91 & 89.02 & 26.50 & 23.93 \\
2 & 2.5 & 34.24 & 23.89 & 61.08 & 54.88 & 88.95 & 26.50 & 23.89 \\
2 & 3.0 & 37.44 & 23.45 & 64.22 & 60.20 & 88.63 & 26.45 & 23.83 \\
2 & 3.5 & 41.16 & 23.01 & 67.43 & 66.22 & 88.29 & 26.45 & 23.70 \\
2 & 4.0 & 44.86 & 22.57 & 71.03 & 72.64 & 87.95 & 26.31 & 23.56 \\
\midrule

3 & 1.2 & 25.91 & 25.28 & 52.36 & 41.45 & 89.84 & 26.40 & 23.90 \\
3 & 1.4 & 26.54 & 25.16 & 53.07 & 42.43 & 89.77 & 26.41 & 23.88 \\
3 & 1.6 & 27.16 & 25.03 & 53.84 & 43.54 & 89.69 & 26.44 & 23.99 \\
3 & 1.8 & 27.92 & 24.92 & 54.47 & 44.45 & 89.63 & 26.43 & 23.97 \\
3 & 2.0 & 28.70 & 24.79 & 55.39 & 45.60 & 89.55 & 26.44 & \textbf{24.06} \\
3 & 2.2 & 29.50 & 24.67 & 56.13 & 46.71 & 89.47 & 26.45 & \underline{24.03} \\
3 & 2.4 & 30.30 & 24.54 & 56.94 & 47.84 & 89.40 & 26.48 & 24.00 \\
3 & 2.5 & 30.64 & 24.47 & 57.53 & 48.55 & 89.34 & 26.49 & 24.00 \\
3 & 3.0 & 33.15 & 24.13 & 59.90 & 51.94 & 89.12 & 26.43 & 23.93 \\
3 & 3.5 & 35.81 & 23.78 & 62.47 & 55.75 & 88.86 & 26.41 & 23.84 \\
3 & 4.0 & 38.62 & 23.44 & 65.10 & 59.95 & 88.60 & 26.36 & 23.75 \\
\midrule

4 & 1.2 & \textbf{24.28} & \textbf{25.62} & \textbf{50.41} & \textbf{38.96} & \textbf{90.02} & 26.37 & 23.92 \\
4 & 1.4 & \underline{24.63} & \underline{25.56} & \underline{50.79} & \underline{39.40} & \underline{89.98} & 26.37 & 23.85 \\
4 & 1.6 & 25.00 & 25.48 & 51.29 & 39.90 & 89.94 & 26.37 & 23.83 \\
4 & 1.8 & 25.33 & 25.41 & 51.70 & 40.43 & 89.89 & 26.41 & 23.95 \\
4 & 2.0 & 25.74 & 25.33 & 52.10 & 40.95 & 89.85 & 26.43 & 23.93 \\
\rowcolor{gray!15}
4 & 2.2 & 26.20 & 25.26 & 52.46 & 41.60 & 89.80 & 26.44 & 23.97 \\
4 & 2.4 & 26.63 & 25.17 & 52.98 & 42.14 & 89.75 & 26.43 & 23.96 \\
4 & 2.5 & 26.87 & 25.13 & 53.42 & 42.54 & 89.71 & 26.44 & 23.97 \\
4 & 3.0 & 28.15 & 24.92 & 55.00 & 44.36 & 89.56 & 26.48 & 23.97 \\
4 & 3.5 & 29.66 & 24.68 & 56.69 & 46.35 & 89.40 & 26.45 & 23.97 \\
4 & 4.0 & 31.30 & 24.47 & 58.47 & 48.44 & 89.22 & 26.42 & 23.98 \\

\bottomrule
\end{tabular}
}
\end{table}
\FloatBarrier

\section{Experimental Results for SD3.5}
\label{sec:exp_sd35}
We further evaluate our method on PIE-Bench under SD3.5 and consider recent frequency-aware editing baselines, including FSI-Edit~\cite{teng2024fsi}, FIA-Edit~\cite{liu2024fia}, and W-EDIT~\cite{zhao2024wedit}. These methods also exploit frequency-domain information or frequency-aware feature interactions in pretrained diffusion models. Since no official implementation of W-EDIT was available to us at the time of our experiments, we restrict the quantitative comparison to FSI-Edit and FIA-Edit. We follow the FlowEdit~\cite{kulikov2024flowedit} sampling configuration and use $(L,\lambda)=(5,2.0)$ for our SD3.5 results unless otherwise specified.

\subsection{Quantitative Results}

As shown in \cref{tab:sd35_quantitative}, we present the quantitative
comparison on PIE-Bench with SD3.5. Our method achieves the best results
on PSNR, LPIPS, MSE, and SSIM, as well as on both Whole and Edited CLIP
similarity. These results indicate that our method is particularly
effective at preserving background appearance during editing, while
maintaining strong semantic alignment in the edited region. In
particular, compared with FSI-Edit and FIA-Edit, our method consistently
improves all background-preservation metrics by a clear margin,
demonstrating the advantage of our frequency-aware design in retaining
low-frequency content and suppressing undesired distortions. Although
FSI-Edit and FIA-Edit attain lower structure distance, this comes at the
cost of weaker semantic editing and reduced background fidelity, which
reflects the same preservation--editability trade-off discussed in the
main text. Moreover, our method retains a competitive inference cost,
indicating that the improved editing quality is not obtained at the price
of a substantially higher runtime. Overall, the quantitative results
verify that our method achieves a more favorable balance among all
compared approaches.

\begin{table}[h]
\caption{
Quantitative comparison on PIE-Bench under SD3.5.
Comparison across all methods without grouping by backbone.
The best and second-best results are shown in \textbf{bold} and
\underline{underline}, respectively. Struct.\ denotes structure distance
and Time denotes the average per-image inference cost.
}
\label{tab:sd35_quantitative}
\centering
\setlength{\tabcolsep}{4pt}
\resizebox{\linewidth}{!}{
\begin{tabular}{lcccccccc}
\toprule
\multirow{2}{*}{Method}
& \multirow{2}{*}{\makecell{Struct.\(\times 10^3\) \\ \(\downarrow\)}}
& \multicolumn{4}{c}{\makecell{Background \\ Preservation}}
& \multicolumn{2}{c}{\makecell{CLIP \\ Similarity}}
& \multirow{2}{*}{\makecell{Time (s) \\ \(\downarrow\)}} \\
\cmidrule(lr){3-6} \cmidrule(lr){7-8}
&
& PSNR \(\uparrow\)
& LPIPS\(\times 10^3\) \(\downarrow\)
& MSE\(\times 10^4\) \(\downarrow\)
& SSIM\(\times 10^2\) \(\uparrow\)
& Whole \(\uparrow\)
& Edited \(\uparrow\)
& \\
\midrule
FSI-Edit~\cite{teng2024fsi}
& \underline{13.64} & 26.63 & 84.43 & 36.42 & 86.33 & 25.68 & 22.42 & 15.57 \\
FIA-Edit~\cite{liu2024fia}
& \textbf{10.34} & \underline{27.32} & \underline{55.03} & \underline{28.66}
& \underline{89.21} & \underline{25.89} & \underline{22.82} & \textbf{5.26} \\
Ours
& 17.41 & \textbf{27.46} & \textbf{41.09} & \textbf{26.46}
& \textbf{91.07} & \textbf{26.43} & \textbf{23.89} & \underline{5.98} \\
\bottomrule
\end{tabular}
}
\end{table}
\subsection{Qualitative Comparisons}
\begin{figure}[!t]
    \centering
    \includegraphics[width=0.95\linewidth]{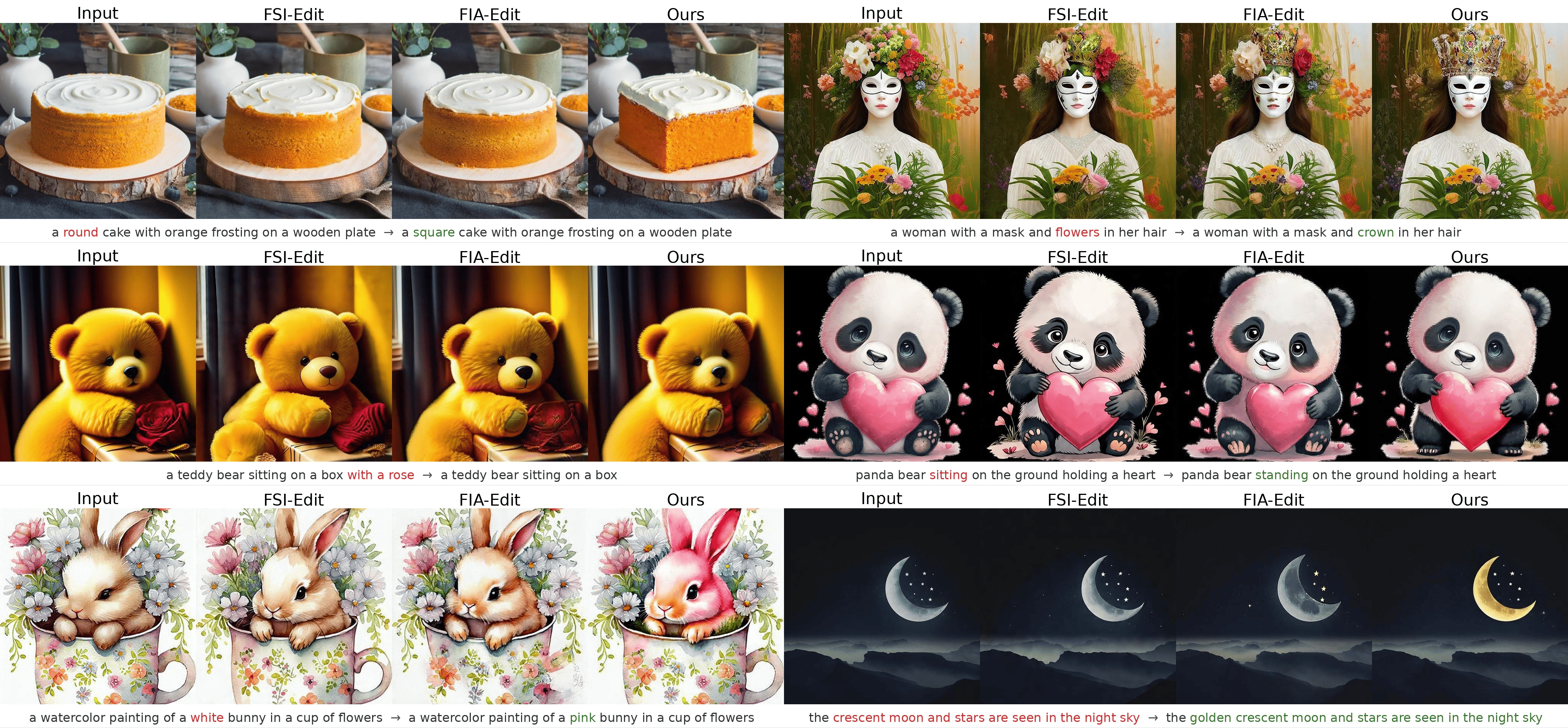}
    \caption{Qualitative comparison on PIE-Bench with SD3.5.}
    \label{fig:qualitative_sd35}
\end{figure}
As shown in \cref{fig:qualitative_sd35}, our method produces edits that
more faithfully match the target prompt while better preserving the
visual structure and identity cues of the input image. Compared with
FSI-Edit and FIA-Edit, it achieves a more favorable balance between
editability and content preservation, realizing the target semantics in
the edited region while keeping unedited regions such as pose, viewpoint,
illumination, and scene layout largely stable.
\subsection{Ablation Study with SD3.5}
Our method introduces two additional hyperparameters: the semantic
strength parameter \(\lambda\) and the wavelet decomposition level \(L\).
To study their effects, we conduct an ablation analysis on PIE-Bench
under SD3.5 by varying \(L\) and \(\lambda\) around the default setting.

As shown in~\cref{tab:ablation_LB_SD35}, increasing \(L\) consistently
improves structural fidelity and background preservation, as reflected by
lower Struct., LPIPS, and MSE together with higher PSNR and SSIM. This
again suggests that higher wavelet decomposition levels are more
effective at preserving low-frequency content. In contrast, increasing
\(\lambda\) enhances semantic editing strength but gradually compromises
structural consistency and background fidelity, while the CLIP similarity
scores remain relatively stable and even peak at smaller \(L\), indicating
a saturation effect in semantic alignment.

Overall, the results show a clear trade-off between preservation quality
and semantic editing performance. Therefore, we adopt
\((L,\lambda)=(5,2.0)\) as the default setting, since it attains the
strongest background preservation while retaining competitive semantic
alignment in the edited region.

\begin{table}[!t]
\caption{
Ablation study of the wavelet decomposition level \(L\) and the semantic
strength parameter \(\lambda\) on PIE-Bench under SD3.5.
The best and second-best results are shown in \textbf{bold} and
\underline{underline}, respectively.
Our final choice is highlighted in gray.
}
\label{tab:ablation_LB_SD35}
\centering
\setlength{\tabcolsep}{4pt}
\resizebox{0.95\linewidth}{!}{
\begin{tabular}{cc c cccc cc}
\toprule
\multirow{2}{*}{\(L\)} & \multirow{2}{*}{\(\lambda\)} & \multirow{2}{*}{\makecell{Struct.\(\times 10^3\) \\ \(\downarrow\)}}
& \multicolumn{4}{c}{\makecell{Background \\ Preservation}}
& \multicolumn{2}{c}{\makecell{CLIP \\ Similarity}} \\
\cmidrule(lr){4-7} \cmidrule(lr){8-9}
& &
& PSNR \(\uparrow\)
& LPIPS\(\times 10^3\) \(\downarrow\)
& MSE\(\times 10^4\) \(\downarrow\)
& SSIM\(\times 10^2\) \(\uparrow\)
& Whole \(\uparrow\)
& Edited \(\uparrow\) \\
\midrule
1 & 1.5 & 22.97 & 26.01 & 47.42 & 35.27 & 90.37 & 26.57 & 23.98 \\
1 & 2.0 & 25.72 & 25.44 & 50.50 & 39.52 & 90.05 & \underline{26.60} & \underline{24.06} \\
1 & 2.5 & 28.77 & 24.88 & 53.72 & 44.20 & 89.71 & 26.59 & 24.03 \\
1 & 3.0 & 31.94 & 24.37 & 56.92 & 49.22 & 89.38 & 26.57 & 23.94 \\
\midrule

2 & 1.5 & 21.68 & 26.33 & 46.01 & 32.80 & 90.55 & 26.57 & 24.03 \\
2 & 2.0 & 23.86 & 25.87 & 48.48 & 36.03 & 90.30 & \textbf{26.61} & 24.05 \\
2 & 2.5 & 26.45 & 25.38 & 51.22 & 39.67 & 90.02 & 26.59 & \textbf{24.07} \\
2 & 3.0 & 29.28 & 24.89 & 54.13 & 43.73 & 89.73 & 26.58 & 24.04 \\
\midrule

3 & 1.5 & 19.87 & 26.86 & 43.87 & 29.71 & 90.79 & 26.47 & 24.00 \\
3 & 2.0 & 21.41 & 26.54 & 45.60 & 31.75 & 90.62 & 26.51 & 24.02 \\
3 & 2.5 & 23.08 & 26.20 & 47.53 & 33.93 & 90.43 & 26.56 & 24.02 \\
3 & 3.0 & 25.09 & 25.82 & 49.77 & 36.56 & 90.21 & 26.55 & 23.99 \\
\midrule

4 & 1.5 & 18.14 & 27.29 & 42.00 & 27.33 & 90.99 & 26.47 & 23.90 \\

4 & 2.0 & 18.99 & 27.12 & 43.05 & 28.29 & 90.88 & 26.47 & 23.93 \\
4 & 2.5 & 20.00 & 26.94 & 44.10 & 29.32 & 90.78 & 26.48 & 23.93 \\
4 & 3.0 & 21.05 & 26.72 & 45.44 & 30.61 & 90.66 & 26.48 & 23.95 \\
\midrule

5 & 1.5 & \textbf{17.06} & \textbf{27.54} & \textbf{40.57} & \textbf{26.06} & \textbf{91.12} & 26.42 & 23.87 \\
\rowcolor{gray!15}
5 & 2.0 & \underline{17.41} & \underline{27.46} & \underline{41.09} & \underline{26.46} & \underline{91.07} & 26.43 & 23.89 \\
5 & 2.5 & 17.77 & 27.40 & 41.56 & 26.78 & 91.02 & 26.44 & 23.90 \\
5 & 3.0 & 18.14 & 27.32 & 42.06 & 27.22 & 90.97 & 26.45 & 23.94 \\
5 & 4.0 & 19.18 & 27.15 & 43.24 & 28.12 & 90.85 & 26.46 & 23.94 \\
\bottomrule
\end{tabular}
}
\end{table}

\section{Additional results for FLUX}
\label{sec:qual_flux}

\begin{figure}[tb]
    \centering
    \includegraphics[width=1.0\linewidth]{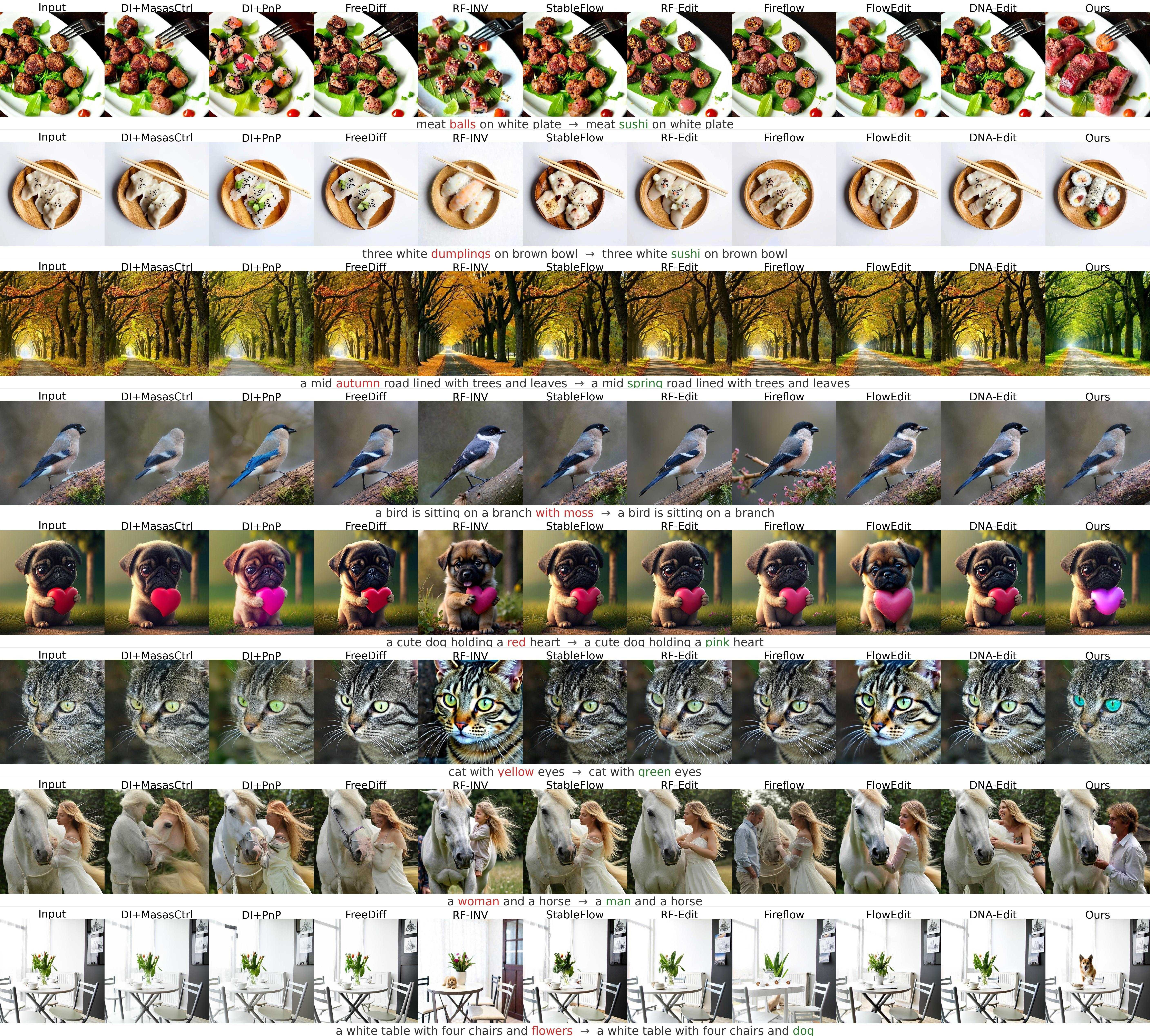}
    \caption{Qualitative comparison on the PIE-Bench with FLUX (Part 1).}
    \label{fig:quali_appen1}
\end{figure}

\begin{figure}[t]
    \centering
    \includegraphics[width=1.0\linewidth]{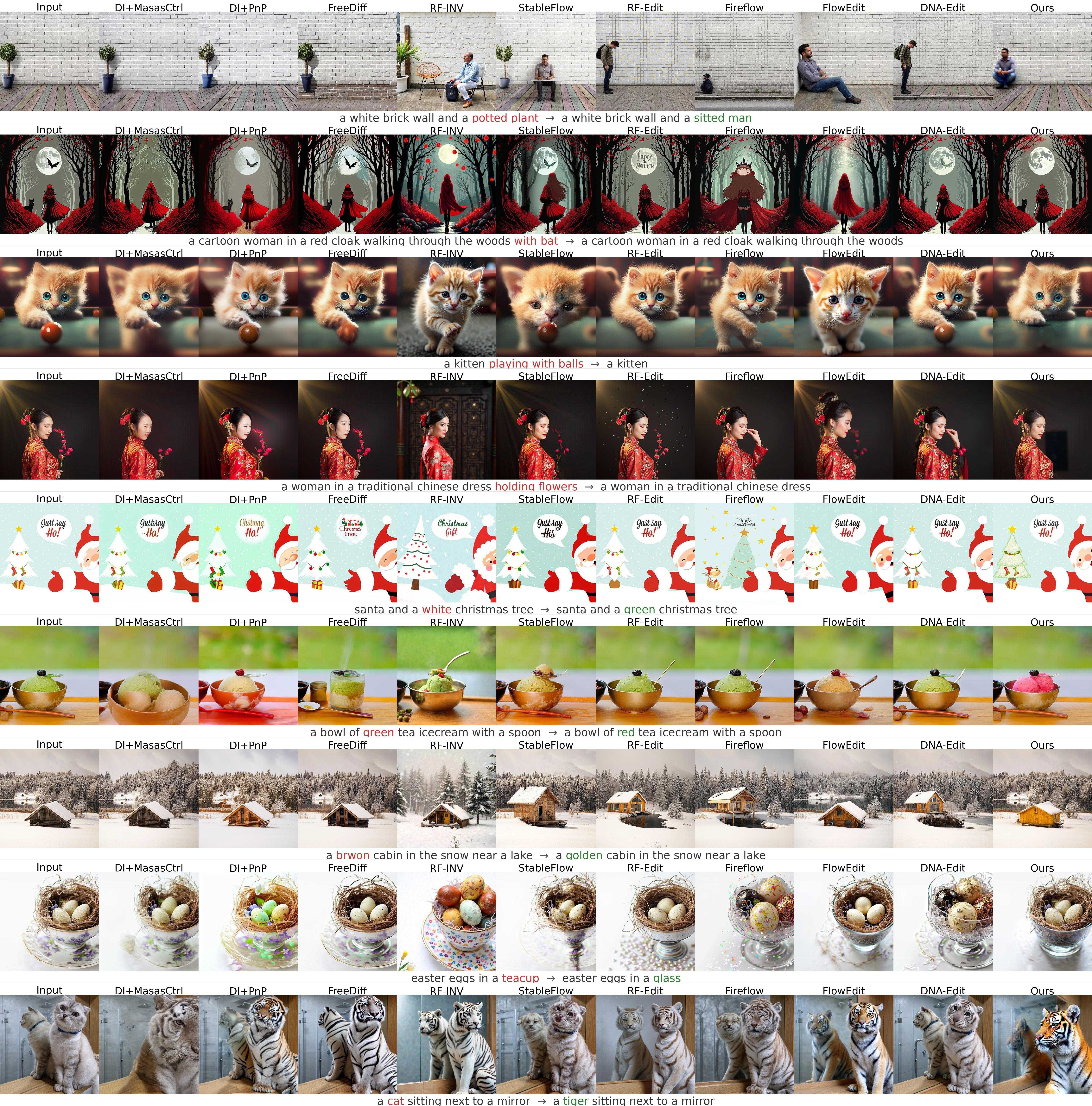}
    \caption{Qualitative comparison on the PIE-Bench with FLUX (Part 2).}
    \label{fig:quali_appen2}
\end{figure}

\begin{figure*}[t]
\centering
\includegraphics[width=\linewidth]{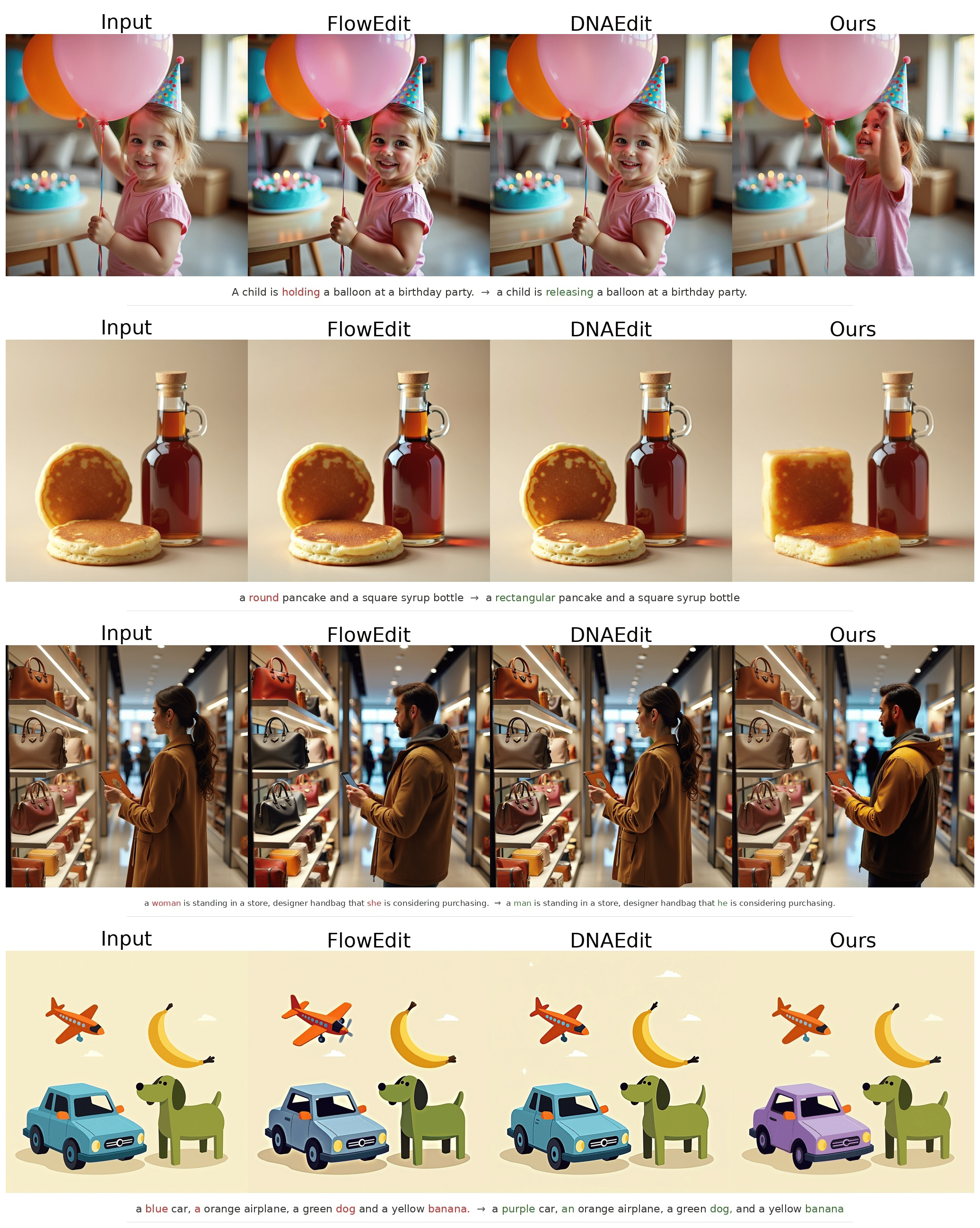}
\caption{
Qualitative comparison on EditBench at \(1024\times1024\) resolution
under the FLUX backbone.
}
\label{fig:editbench_qualitative}
\end{figure*}

In this section, we first present additional PIE-Bench qualitative results under the FLUX backbone in~\cref{fig:quali_appen1,fig:quali_appen2}. To further evaluate the generalization ability of our method beyond PIE-Bench, we conduct additionally high-resolution experiments on EditBench at \(1024\times1024\) resolution using 200 samples. All results in this section are obtained under the FLUX backbone. As reported in~\cref{tab:editbench}, our method consistently outperforms FlowEdit on all metrics and also achieves the best results on most background preservation and CLIP similarity metrics. Compared with FlowEdit, our method substantially improves PSNR, LPIPS, MSE, and SSIM, indicating better preservation of unedited regions at high resolution. It also achieves higher Whole and Edited CLIP similarities, showing stronger consistency with the target prompts. Compared with DNA-Edit, our method obtains higher PSNR, lower LPIPS, higher SSIM, and stronger CLIP scores, while DNA-Edit achieves the lowest Struct. and MSE. These results indicate that our method maintains strong background fidelity while producing more semantically aligned edits. For the high-resolution EditBench setting, we also examine the resolution dependence of the wavelet level \(L\). While the default FLUX setting in the main PIE-Bench experiments uses \(L=3\) at \(512\times512\) resolution, we further evaluate \(L \in \{2,3,4\}\) at \(1024\times1024\) resolution. The results show that \(L=3\) remains a strong choice at high resolution, achieving the best PSNR of 32.30 and the highest Edited CLIP score of 22.88, compared with 30.30/22.86 for \(L=2\) and 31.66/22.73 for \(L=4\). Therefore, we use \(L=3\) for the EditBench evaluation. As shown in~\cref{fig:editbench_qualitative}, our method produces edits that better follow the target prompts while keeping the unedited regions visually stable. These qualitative results are consistent with the quantitative improvements reported in~\cref{tab:editbench}.
\begin{table}[h]
\caption{
Quantitative comparison on EditBench at \(1024\times1024\) resolution
(200 samples) under the FLUX backbone. The best and second-best results
are shown in \textbf{bold} and \underline{underline}, respectively.
Struct.\ denotes structure distance.
}
\label{tab:editbench}
\centering
\setlength{\tabcolsep}{5pt}
\resizebox{0.95\linewidth}{!}{
\begin{tabular}{lccccccc}
\toprule
\multirow{2}{*}{Method}
& \multirow{2}{*}{\makecell{Struct.\(\times 10^3\) \\ \(\downarrow\)}}
& \multicolumn{4}{c}{\makecell{Background \\ Preservation}}
& \multicolumn{2}{c}{\makecell{CLIP \\ Similarity}} \\
\cmidrule(lr){3-6} \cmidrule(lr){7-8}
&
& PSNR \(\uparrow\)
& LPIPS\(\times 10^3\) \(\downarrow\)
& MSE\(\times 10^4\) \(\downarrow\)
& SSIM\(\times 10^2\) \(\uparrow\)
& Whole \(\uparrow\)
& Edited \(\uparrow\) \\
\midrule
FlowEdit~\cite{kulikov2024flowedit} & 8.32 & 26.10 & 39.23 & 30.14 & 94.88 & \underline{27.03} & \underline{21.99} \\
DNA-Edit~\cite{xie2025dnaedit}
& \textbf{2.95} & \underline{31.84} & \underline{23.86} & \textbf{8.19}
& \underline{97.28} & 26.68 & 21.39 \\
Ours
& \underline{7.43} & \textbf{32.30} & \textbf{12.22} & \underline{9.16}
& \textbf{98.24} & \textbf{27.96} & \textbf{22.88} \\
\bottomrule
\end{tabular}
}
\end{table}

% \clearpage
\FloatBarrier

\section{Ablation Study with FLUX}
\label{sec:FLUX_parameter}
In this section, we provide a comprehensive ablation analysis of our method under the FLUX setting on PIE-Bench. 
Specifically, we study three aspects of the design: the effects of the two key hyperparameters, namely the semantic strength parameter $\lambda$ and the wavelet decomposition level $L$; the role of frequency-band selection and the choice of low-pass filter in the compensation module; and the impact of different time-dependent weight injection strategies. 
Together, these experiments are intended to examine both the performance sensitivity of the method and the design choices adopted in the main implementation.
\subsection{Ablation Study of Hyperparameters}
\begin{figure}[t]
    \centering
    \includegraphics[width=0.95\linewidth]{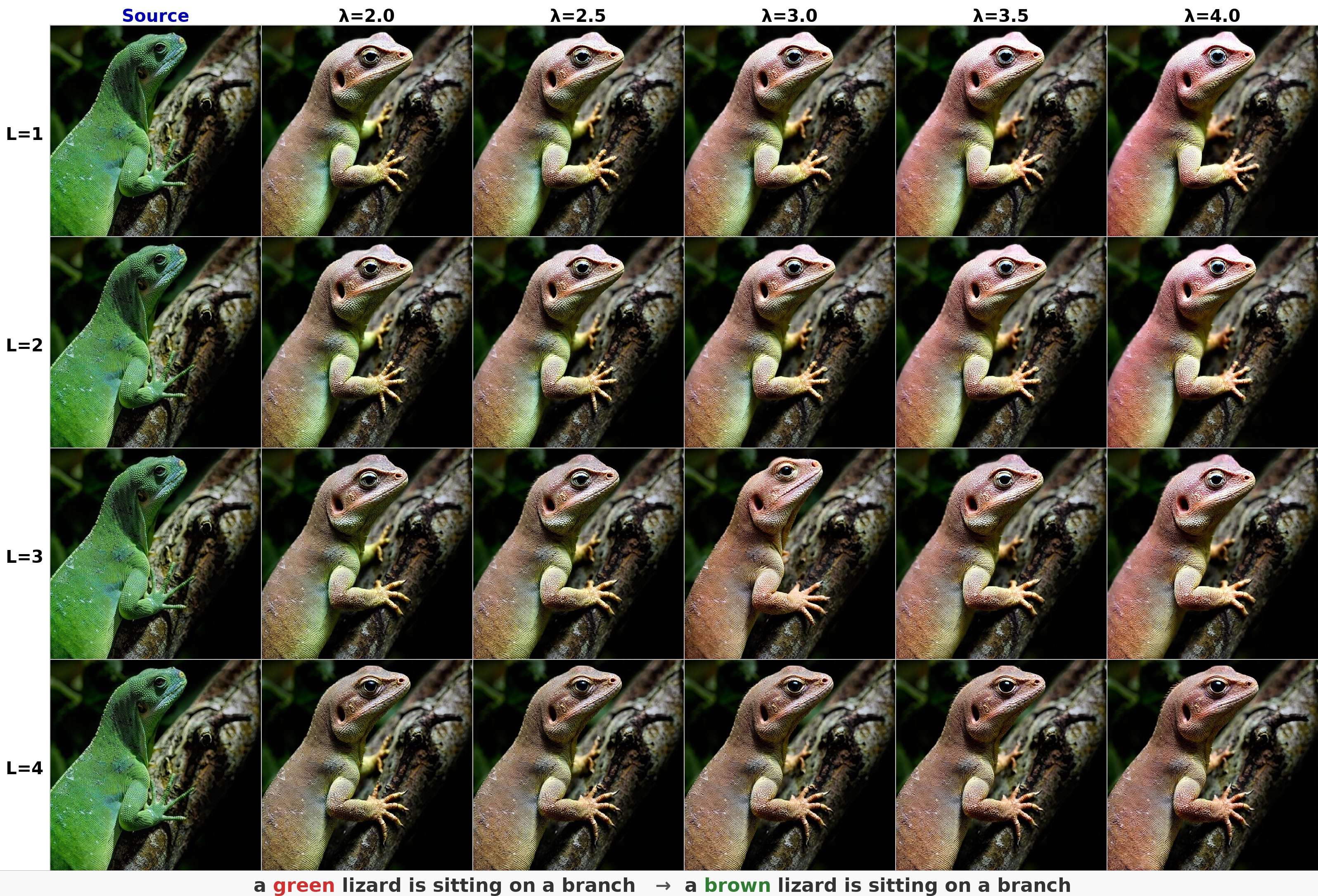}
    \caption{Qualitative ablation on $L$ and $\lambda$ under FLUX.}
\label{fig:ablation_FLUX_vis}
\end{figure}

Our method introduces two additional hyperparameters: the semantic strength parameter $\lambda$ and the wavelet decomposition level $L$. The main text uses $L=3$ and $\lambda=3.0$ as the default setting. To further examine the influence of $L$ and $\lambda$, we report the ablation results in~\cref{tab:ablation_LB}. Overall, increasing $L$ tends to improve structural and background preservation, as reflected by lower Struct., LPIPS, and MSE, together with higher PSNR and SSIM. In contrast, the CLIP similarity scores do not monotonically improve with larger $L$, and the best semantic alignment is achieved at moderate values of the two hyperparameters. At the same time, the results remain relatively stable within a nearby range around the default setting, with only small differences across configurations. Considering the overall trade-off between preservation quality and semantic editing performance, we use $L=3$ and $\lambda=3.0$ as the default setting in the main text. As shown in~\cref{fig:ablation_FLUX_vis}, the qualitative results exhibit a similar trend, where increasing $\lambda$ generally leads to stronger edits, while increasing $L$ improves structural preservation. In this example, the edited lizard also shows a more consistent brown appearance with less residual green color across this range of configurations.

\begin{table}[h]
\renewcommand{\arraystretch}{1.08}
\caption{%
  Ablation study on the hyperparameters $L$ and $\lambda$ on PIE-Bench under FLUX\@.
  \textbf{Bold} and \underline{underline} denote the best and second-best results
  per metric across all configurations, respectively.%
}
\label{tab:ablation_LB}
\centering
\setlength{\tabcolsep}{6pt}
\resizebox{0.8\linewidth}{!}{%
\begin{tabular}{cc c cccc cc}
\toprule
\multirow{2}{*}{$L$}
  & \multirow{2}{*}{$\lambda$}
  & \multirow{2}{*}{\makecell{Struct.$\times10^{3}$\\$\downarrow$}}
  & \multicolumn{4}{c}{\makecell{Background\\Preservation}}
  & \multicolumn{2}{c}{\makecell{CLIP\\Similarity}} \\
\cmidrule(lr){4-7}\cmidrule(lr){8-9}
  & &
  & PSNR\,$\uparrow$
  & LPIPS$\times10^{3}$\,$\downarrow$
  & MSE$\times10^{4}$\,$\downarrow$
  & SSIM$\times10^{2}$\,$\uparrow$
  & Whole\,$\uparrow$
  & Edited\,$\uparrow$ \\
\midrule

1 & 2.0 & 28.81 & 26.12 & 53.02 & 38.98 & 90.66 & 25.52 & 23.08 \\
1 & 2.5 & 31.93 & 25.68 & 55.75 & 42.46 & 90.31 & 25.53 & 23.06 \\
1 & 2.7 & 33.34 & 25.51 & 56.92 & 43.97 & 90.16 & \underline{25.58} & 23.06 \\
1 & 2.9 & 34.66 & 25.34 & 58.04 & 45.44 & 90.01 & \textbf{25.59} & 23.02 \\
1 & 3.0 & 35.31 & 25.26 & 58.48 & 46.12 & 89.95 & 25.56 & 23.01 \\
1 & 3.2 & 36.69 & 25.09 & 59.70 & 47.76 & 89.80 & \textbf{25.59} & 23.00 \\
1 & 3.3 & 37.40 & 25.00 & 60.33 & 48.59 & 89.72 & 25.57 & 22.95 \\
1 & 3.5 & 38.70 & 24.85 & 61.45 & 50.18 & 89.57 & 25.56 & 22.92 \\
1 & 4.0 & 42.05 & 24.45 & 64.22 & 54.21 & 89.20 & 25.51 & 22.83 \\
\midrule

2 & 2.0 & 27.82 & 26.35 & 52.10 & 37.47 & 90.82 & 25.54 & \textbf{23.10} \\
2 & 2.5 & 31.01 & 25.93 & 54.58 & 40.62 & 90.50 & 25.55 & 23.07 \\
2 & 2.7 & 32.21 & 25.76 & 55.71 & 42.07 & 90.36 & \underline{25.58} & 23.02 \\
2 & 2.9 & 33.57 & 25.59 & 56.75 & 43.44 & 90.23 & \underline{25.58} & 23.02 \\
2 & 3.0 & 34.21 & 25.51 & 57.36 & 44.14 & 90.16 & \textbf{25.59} & 23.01 \\
2 & 3.2 & 35.54 & 25.35 & 58.43 & 45.60 & 90.03 & 25.56 & 22.98 \\
2 & 3.3 & 36.21 & 25.26 & 58.95 & 46.36 & 89.96 & 25.55 & 22.98 \\
2 & 3.5 & 37.49 & 25.10 & 60.00 & 47.87 & 89.82 & 25.54 & 23.01 \\
2 & 4.0 & 40.84 & 24.71 & 62.80 & 51.81 & 89.47 & 25.46 & 22.85 \\
\midrule

3 & 2.0 & 25.72 & 26.88 & 49.64 & 34.09 & 91.16 & 25.48 & 22.98 \\
3 & 2.5 & 28.30 & 26.52 & 51.83 & 36.59 & 90.88 & 25.54 & 23.05 \\
3 & 2.7 & 29.35 & 26.38 & 52.72 & 37.62 & 90.77 & 25.53 & 23.07 \\
3 & 2.9 & 30.38 & 26.25 & 53.49 & 38.57 & 90.67 & 25.51 & 23.05 \\
\rowcolor{gray!15}
3 & 3.0 & 30.95 & 26.17 & 54.02 & 39.24 & 90.60 & 25.52 & 23.05 \\
3 & 3.2 & 31.98 & 26.02 & 54.96 & 40.32 & 90.48 & 25.53 & 23.08 \\
3 & 3.3 & 32.56 & 25.95 & 55.41 & 40.89 & 90.43 & 25.53 & 23.05 \\
3 & 3.5 & 33.67 & 25.80 & 56.24 & 41.99 & 90.33 & 25.55 & \underline{23.09} \\
3 & 4.0 & 36.65 & 25.44 & 58.68 & 45.00 & 90.04 & 25.44 & 22.94 \\
\midrule

4 & \textbf{2.0} & \textbf{22.37} & \textbf{27.55} & \textbf{46.17} & \textbf{30.18} & \textbf{91.56} & 25.38 & 22.90 \\
4 & \underline{2.5} & \underline{24.14} & \underline{27.30} & \underline{47.61} & \underline{31.71} & \underline{91.37} & 25.46 & 22.88 \\
4 & 2.7 & 24.87 & 27.20 & 48.19 & 32.36 & 91.30 & 25.46 & 22.91 \\
4 & 2.9 & 25.68 & 27.10 & 48.79 & 33.00 & 91.22 & 25.40 & 22.90 \\
4 & 3.0 & 26.06 & 27.05 & 49.12 & 33.36 & 91.17 & 25.39 & 22.90 \\
4 & 3.2 & 26.85 & 26.95 & 49.73 & 33.98 & 91.10 & 25.39 & 22.90 \\
4 & 3.3 & 27.24 & 26.90 & 50.02 & 34.28 & 91.06 & 25.40 & 22.88 \\
4 & 3.5 & 28.13 & 26.80 & 50.69 & 35.10 & 90.98 & 25.40 & 22.87 \\
4 & 4.0 & 30.31 & 26.54 & 52.51 & 36.97 & 90.77 & 25.42 & 22.86 \\
\bottomrule
\end{tabular}%
}
\end{table}

\subsection{Ablation on Frequency-Band Selection and Filter Choice}
\label{sec:freq_ablation}

A key premise of our method is that editing global attributes such as color and material requires the trajectory to establish a coherent semantic shift at an early stage. If such a shift is not formed early, later updates tend to remain limited to local refinements, which weakens the desired semantic transfer. Based on this motivation, we study whether injecting low-frequency information from the semantic editing signal $\vsem$ can provide a more effective compensation signal. We further examine whether the Haar wavelet is a suitable default choice among different low-pass operators. The results are summarized in~\cref{tab:filter_ablation}.

\paragraph{Implementation details}
Throughout this ablation study, we fix the Haar wavelet decomposition level to $L=3$. This corresponds to a three-level decomposition that retains the coarsest approximation component. For a fair comparison in Group II, all alternative low-pass operators are configured to have a comparable effective cutoff range. Specifically, the Gaussian filter uses a standard deviation of $\sigma=2^{L-1}=4$ in the latent space, with kernel size $k=\lceil 6\sigma+1\rceil=25$~\cite{jahne2005digital,tschirsich2015notes}. The Butterworth filter~\cite{butterworth1930theory,jahne2005digital} is implemented in the Fourier domain with cutoff radii $r_h=H/2^{L+1}$ and $r_w=W/2^{L+1}$, and filter order $n=2$, with transfer function:
\begin{equation}
    H(f_h,f_w)=\frac{1}{1+\left(\left(\frac{f_h}{r_h}\right)^2+\left(\frac{f_w}{r_w}\right)^2\right)^n}.
\end{equation}
The ideal low-pass filter~\cite{jahne2005digital} uses the same cutoff radii with a binary mask in the Fourier domain,
\begin{equation}
    H(f_h,f_w)=\mathbf{1}\!\left[\left(\frac{f_h}{r_h}\right)^2+\left(\frac{f_w}{r_w}\right)^2\le 1\right].
\end{equation}
We also include average pooling as a simple spatial smoothing baseline. The latent feature is partitioned into non-overlapping blocks of size $2^L\times 2^L$, each block is replaced by its mean, and the result is then upsampled to the original resolution using bilinear interpolation.

\paragraph{Group I: Frequency-band selection}
We compare four variants that differ in the signal used for compensation: zero, which disables compensation and reduces to the original FlowEdit pipeline; identity, which directly uses the full-spectrum semantic editing signal $\vsem$; highpass, which keeps only the detail sub-bands; and wavelet, which keeps only the low-frequency approximation component.

As shown in~\cref{tab:filter_ablation}, removing compensation leads to substantially worse background preservation metrics, although it attains the lowest Struct. score. Using the full-spectrum signal or only the high-frequency component improves some metrics, but neither provides the best overall trade-off. By comparison, the low-frequency wavelet variant achieves the best results on most metrics and the second-best result on Struct. These results support the benefit of compensation and suggest that the low-frequency component provides the most suitable signal for this purpose.

\paragraph{Group II: Low-pass filter choice}
We further compare the Haar wavelet with several alternative low-pass operators under a comparable effective cutoff range. As shown in~\cref{tab:filter_ablation}, all low-pass variants fall within a relatively similar performance range, which suggests that the low-frequency component of the compensation signal is the main factor. At the same time, the Haar wavelet gives the strongest overall results across the table. This indicates that the benefit is not only from suppressing high-frequency content, but also from using a low-frequency extraction operator that is well suited to the compensation objective. Based on these observations, we use the Haar wavelet as the default operator. It provides strong overall performance and introduces only a simple and interpretable hyperparameter $L$.

\begin{table}[h]
\caption{
Ablation study on frequency-band selection and low-pass filter
choice on the PIE-Bench.
Group~I examines which frequency band of the semantic editing signal $\vsem$
should be used for compensation: none (Zero, equivalent to the
original FlowEdit), full spectrum (Identity), high-frequency
sub-bands only (Highpass), and low-frequency approximation only
(Haar wavelet, ours).
Group~II compares alternative low-pass filters with matched
effective bandwidth against the Haar wavelet.
All variants use FLUX as the backbone.
The best and second-best results are shown in \textbf{bold}
and \underline{underline}, respectively.
}
\label{tab:filter_ablation}
\centering
\setlength{\tabcolsep}{4.6pt}
\resizebox{0.9\textwidth}{!}{%
\begin{tabular}{lccccccc}
\toprule
\multirow{2}{*}{Variant}
& \multirow{2}{*}{\makecell{Struct.$\times 10^3$ \\ $\downarrow$}}
& \multicolumn{4}{c}{Background Preservation}
& \multicolumn{2}{c}{CLIP Similarity} \\
\cmidrule(lr){3-6} \cmidrule(lr){7-8}
&
& PSNR $\uparrow$
& LPIPS$\times 10^3$ $\downarrow$
& MSE$\times 10^4$ $\downarrow$
& SSIM$\times 10^2$ $\uparrow$
& Whole $\uparrow$
& Edited $\uparrow$ \\
\midrule
\multicolumn{8}{l}{\emph{Group I: Frequency-band selection}} \\[2pt]
Zero (no compensation)    & \textbf{27.61} & 22.23          & 111.32         & 90.98          & 83.64          & \underline{25.39} & 22.62          \\
Identity (full spectrum)  & 41.53          & 23.61          & 69.77          & 63.31          & 88.05          & 25.03             & 22.24          \\
Highpass only             & 33.20          & 24.81          & 59.69          & 50.37          & 89.25          & 24.97             & 22.34          \\
Haar wavelet (Ours)       & \underline{30.95} & \textbf{26.17}    & \textbf{54.02}    & \textbf{39.24}    & \textbf{90.60}    & \textbf{25.52}    & \textbf{23.05}    \\
\midrule
\multicolumn{8}{l}{\emph{Group II: Low-pass filter comparison}} \\[2pt]
Average Pooling           & 33.01             & 25.51             & 57.41             & 43.85             & 90.00             & 25.22             & 22.75             \\
Gaussian                  & 31.42             & \underline{25.65} & \underline{56.61} & \underline{42.68} & \underline{90.11} & 25.27             & \underline{22.87} \\
Butterworth               & 35.72             & 25.05             & 60.48             & 48.08             & 89.61             & 25.23             & 22.74             \\
Ideal LP (FFT)            & 37.48             & 24.87             & 61.98             & 50.27             & 89.44             & 25.28             & 22.66             \\
Haar wavelet (Ours)       & \underline{30.95} & \textbf{26.17}    & \textbf{54.02}    & \textbf{39.24}    & \textbf{90.60}    & \textbf{25.52}    & \textbf{23.05}    \\
\bottomrule
\end{tabular}
}
\end{table}

\FloatBarrier

\subsection{Ablation Study of Time-Dependent Weight Injection}
\label{sec:time_ablation}

\cref{tab:ablation_flux_time} reports a simple ablation on different time-dependent weight injection strategies under FLUX, with $\lambda=3.0$ fixed. 
Overall, the different variants produce visually similar edited results, suggesting that the choice of $w(t)$ does not substantially affect the perceptual appearance in this example. 
Quantitatively, placing relatively larger weights on earlier, higher-noise timesteps (equivalently, suppressing the low-noise timesteps more strongly) generally improves structural consistency and background preservation, as reflected by the progressive gains from $\lambda$ to $\lambda t^3$.
Among these variants, $\lambda t^3$ achieves the best preservation-oriented results. 
However, its advantage over $\lambda t^2$ remains relatively small, and the two settings are close in terms of semantic alignment. 
Considering the overall trade-off, $\lambda t^2$ already provides a favorable balance between strong background preservation and effective semantic editing. 
Therefore, we adopt $\lambda t^2$ as the default setting in the main implementation.

\begin{table}[h]
\caption{
Ablation study of the proposed method on \textbf{FLUX}.
We vary the time-dependent weight injection method $w(t)$,
while keeping $\lambda = 3.0$ fixed.
}
\label{tab:ablation_flux_time}
\centering
\setlength{\tabcolsep}{5pt}
\resizebox{0.9\linewidth}{!}{
\begin{tabular}{c ccccccc}
\toprule
\multirow{2}{*}{$w(t)$} &
\multirow{2}{*}{\makecell{Struct.$\times 10^3$ \\ $\downarrow$}} &
\multicolumn{4}{c}{\makecell{Background \\ Preservation}} &
\multicolumn{2}{c}{\makecell{CLIP \\ Similarity}} \\
\cmidrule(lr){3-6} \cmidrule(lr){7-8}

&
& PSNR $\uparrow$
& LPIPS$\times10^3$ $\downarrow$
& MSE$\times10^4$ $\downarrow$
& SSIM$\times10^2$ $\uparrow$
& Whole $\uparrow$
& Edited $\uparrow$ \\

\midrule

$\lambda$ & 35.88 & 25.38 & 60.33 & 45.56 & 89.81 & 25.49 & 22.93 \\
$\lambda t$ & 33.02 & 25.83 & 56.40 & 41.79 & 90.30 & 25.46 & 23.03 \\
$\lambda t^2$ & 30.95 & 26.17 & 54.02 & 39.24 & 90.60 & 25.52 & 23.05 \\

$\lambda t^3$ & 29.18 & 26.43 & 52.28 & 37.28 & 90.83 & 25.41 & 23.01 \\
Exp &31.82 & 26.02 & 55.10 & 40.33 & 90.46 & 25.08 & 22.96\\
\bottomrule
\end{tabular}
}
\end{table}

\section{Runtime Analysis}
\label{sec:runtime}
We report the average per-image inference time of our method and several
representative baselines on PIE-Bench. All measurements are obtained on a
single NVIDIA A800 (80GB) GPU and averaged over 100 samples. As
summarized in \cref{tab:runtime}, our method runs at a speed comparable
to inversion-free baselines such as FlowEdit and DNA-Edit, while being
markedly faster than inversion-based approaches such as RF-Edit and
FireFlow. This indicates that the proposed wavelet-guided compensation
introduces only a modest computational overhead on top of the
inversion-free editing pipeline, while delivering the consistent quality
gains reported above.

\begin{table}[h]
\caption{
Average per-image inference time (seconds) on PIE-Bench, measured on a
single NVIDIA A800 (80GB) GPU and averaged over 100 samples.
}
\label{tab:runtime}
\centering
\setlength{\tabcolsep}{6pt}
\resizebox{0.80\linewidth}{!}{
\begin{tabular}{lcccccc}
\toprule
Method & RF-Inv & RF-Edit & FireFlow & FlowEdit & DNA-Edit & Ours \\
\midrule
Time (s) \(\downarrow\) & 9.46 & 15.60 & 15.83 & 7.75 & 7.80 & 10.11 \\
\bottomrule
\end{tabular}
}
\end{table}

\section{User Study Details}
\label{sec:user}
To provide additional details of the user study described in the main manuscript, we present the evaluation interface and the complete preference statistics in this section.

\begin{figure}[htbp]
    \centering
    \includegraphics[width=0.95\linewidth]{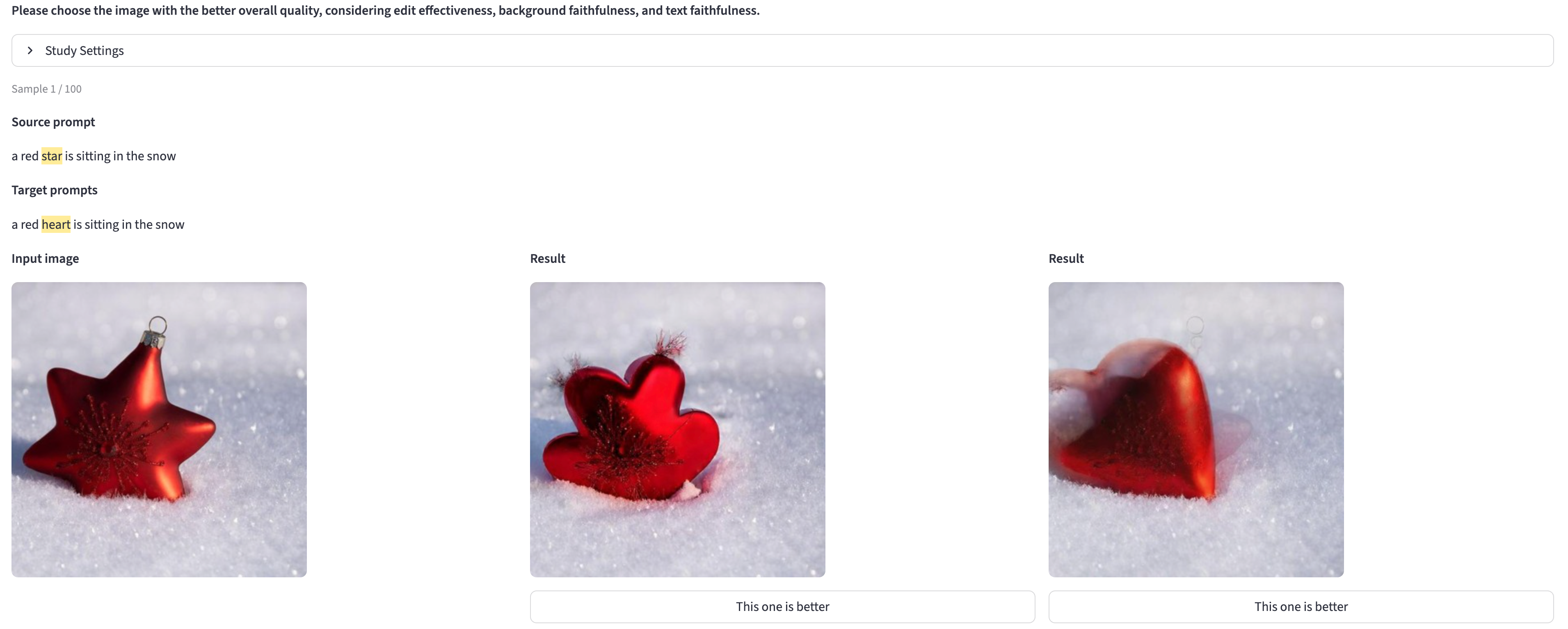}
    \caption{Screenshot of the user study.}
    \label{fig:user_interface}
\end{figure}

We implement a web-based interface to collect human feedback. As shown in \cref{fig:user_interface}, each evaluation instance displays the source prompt and the target prompt, where the modified words are highlighted in yellow to help evaluators quickly identify the intended edit.

Below the prompts, the interface presents the input image together with two edited results: One generated by our method and the other by a baseline. The spatial order of the two images is randomly shuffled in each trial to avoid position bias. Evaluators select their preferred result by clicking the button below the corresponding image.

The study involves 10 independent participants. For each baseline comparison, we randomly sample 100 image pairs from the PIE-Bench dataset. Participants are blind to the identity of the compared methods. The aggregated preference statistics are reported in \cref{tab:user_study_appendix}, where our method consistently achieves higher preference rates across all baselines.

\begin{table}[tbp]
\centering
\small
\caption{User preference rate (\%) on PIE-Bench under pairwise comparison.}
\label{tab:user_study_appendix}

\begin{minipage}[t]{0.27\linewidth}
\centering
\setlength{\tabcolsep}{3pt}
\begin{tabular}{lcc}
\toprule
Method & Ours & Base \\
\midrule
MasaCtrl & 78 & 22 \\
PnP      & 68 & 32 \\
FreeDiff & 58 & 42 \\
\bottomrule
\end{tabular}
\\[2pt]
(a) Diffusion-based methods
\end{minipage}
\hspace{\fill}
\begin{minipage}[t]{0.27\linewidth}
\centering
\setlength{\tabcolsep}{3pt}
\begin{tabular}{lcc}
\toprule
Method & Ours & Base \\
\midrule
DVRF      & 58 & 42 \\
FlowEdit  & 56 & 44 \\
FlowAlign & 54 & 46 \\
\bottomrule
\end{tabular}
\\[2pt]
(b) SD3-based methods
\end{minipage}
\hspace{\fill}
\begin{minipage}[t]{0.27\linewidth}
\centering
\setlength{\tabcolsep}{3pt}
\begin{tabular}{lcc}
\toprule
Method & Ours & Base \\
\midrule
FireFlow   & 71 & 29 \\
StableFlow & 68 & 32 \\
FlowEdit   & 66 & 34 \\
RF-Inv     & 64 & 36 \\
DNA-Edit   & 58 & 42 \\
RF-Edit    & 58 & 42 \\
\bottomrule
\end{tabular}
\\[2pt]
(c) FLUX-based methods
\end{minipage}
\end{table}

\clearpage
\bibliographystyle{IEEEtran}
\bibliography{main}
\end{document}

%% file: preamble.tex
\usepackage{floatflt}
\usepackage{algorithm}
\usepackage{algpseudocode}
\usepackage[capitalize]{cleveref}
\usepackage{graphicx}
\usepackage{tabularx}
\usepackage{wrapfig}
\usepackage{multirow}
\usepackage{makecell}
\usepackage{enumitem}

\usepackage{amsmath}
\Crefname{equation}{Eq.}{Eqs.} % 大写形式（用于 \Cref）
\crefname{equation}{Eq.}{Eqs.} % 小写形式（用于 \cref）

\definecolor{forth}{RGB}{255,255,255}  % 浅粉
\definecolor{third}{RGB}{255,240,192} % 中粉
\definecolor{second}{RGB}{192,216,255}   % 深粉
\definecolor{best}{RGB}{255,192,192}
\definecolor{lightgray}{RGB}{180, 180, 180}

\usepackage[mathscr]{eucal}
\usepackage{epsfig,epsf,psfrag}
\usepackage{amssymb,amsmath,amsfonts,latexsym}
\usepackage{amsmath,graphicx,bm,xcolor,url}
\usepackage{fixltx2e}%ordering of single and double column floats
\usepackage{array}%array and tabular environments
\usepackage{verbatim}
\usepackage{bm}
\usepackage{verbatim}
\usepackage{textcomp}
\usepackage{mathrsfs}
\usepackage{epstopdf}
\usepackage{relsize}
\catcode`~=11 \def\UrlSpecials{\do\~{\kern -.15em\lower .7ex\hbox{~}\kern .04em}} \catcode`~=13 

\allowdisplaybreaks[3]

\DeclareMathAlphabet{\mathbsf}{OT1}{cmss}{bx}{n}
\DeclareMathAlphabet{\mathssf}{OT1}{cmss}{m}{sl}% slanted sans serif

\DeclareSymbolFont{bsfletters}{OT1}{cmss}{bx}{n}  
\DeclareSymbolFont{ssfletters}{OT1}{cmss}{m}{n}
\DeclareMathSymbol{\bsfGamma}{0}{bsfletters}{'000}
\DeclareMathSymbol{\ssfGamma}{0}{ssfletters}{'000}
\DeclareMathSymbol{\bsfDelta}{0}{bsfletters}{'001}
\DeclareMathSymbol{\ssfDelta}{0}{ssfletters}{'001}
\DeclareMathSymbol{\bsfTheta}{0}{bsfletters}{'002}
\DeclareMathSymbol{\ssfTheta}{0}{ssfletters}{'002}
\DeclareMathSymbol{\bsfLambda}{0}{bsfletters}{'003}
\DeclareMathSymbol{\ssfLambda}{0}{ssfletters}{'003}
\DeclareMathSymbol{\bsfXi}{0}{bsfletters}{'004}
\DeclareMathSymbol{\ssfXi}{0}{ssfletters}{'004}
\DeclareMathSymbol{\bsfPi}{0}{bsfletters}{'005}
\DeclareMathSymbol{\ssfPi}{0}{ssfletters}{'005}
\DeclareMathSymbol{\bsfSigma}{0}{bsfletters}{'006}
\DeclareMathSymbol{\ssfSigma}{0}{ssfletters}{'006}
\DeclareMathSymbol{\bsfUpsilon}{0}{bsfletters}{'007}
\DeclareMathSymbol{\ssfUpsilon}{0}{ssfletters}{'007}
\DeclareMathSymbol{\bsfPhi}{0}{bsfletters}{'010}
\DeclareMathSymbol{\ssfPhi}{0}{ssfletters}{'010}
\DeclareMathSymbol{\bsfPsi}{0}{bsfletters}{'011}
\DeclareMathSymbol{\ssfPsi}{0}{ssfletters}{'011}
\DeclareMathSymbol{\bsfOmega}{0}{bsfletters}{'012}
\DeclareMathSymbol{\ssfOmega}{0}{ssfletters}{'012}

\newcommand{\qednew}{\nobreak \ifvmode \relax \else
      \ifdim\lastskip<1.5em \hskip-\lastskip
      \hskip1.5em plus0em minus0.5em \fi \nobreak
      \vrule height0.75em width0.5em depth0.25em\fi}

\newcommand{\bv}{\bm{v}}
\newcommand{\bx}{\bm{x}}

\newcommand{\be}{\bm{e}}
\newcommand{\btheta}{\bm{\theta}}

\newcommand{\xsrc}{\bm{x}^{\mathrm{src}}}
\newcommand{\xsrci}{\bm{x}_{t_i}^{\mathrm{src}}}
\newcommand{\xtari}{\bm{x}_{t_i}^{\mathrm{tar}}}

\newcommand{\esrc}{\bm{e}_{\mathrm{src}}}
\newcommand{\etar}{\bm{e}_{\mathrm{tar}}}
\newcommand{\vtheta}{\bm{v}_{\btheta}}
\newcommand{\Dv}{\Delta\bm{v}}

\newcommand{\vsem}{\bm{v}_{\mathrm{sem}}}
\newcommand{\vlow}{\bm{v}_{\mathrm{low}}}
\newcommand{\LF}{\mathcal{F}_{\!\mathrm{low}}}
\newcommand{\DWT}{\mathrm{DWT}}
\newcommand{\IDWT}{\mathrm{IDWT}}
\newcommand{\RR}{\mathbb{R}}

%% file: main.bib
@String(CVPR  = {IEEE Conf. Comput. Vis. Pattern Recog.})

@String(ICCV  = {Int. Conf. Comput. Vis.})

@String(ECCV  = {Eur. Conf. Comput. Vis.})

@String(NeurIPS = {Adv. Neural Inform. Process. Syst.})

@String(ICML  = {Int. Conf. Mach. Learn.})

@String(ICLR  = {Int. Conf. Learn. Represent.})

@String(AAAI  = {AAAI})

@String(CVPR  = {CVPR})

@String(ICCV  = {ICCV})

@String(ECCV  = {ECCV})

@String(NeurIPS = {NeurIPS})

@String(ICML  = {ICML})

@String(ICLR  = {ICLR})

@inproceedings{wallace2023edict,
  title={{EDICT}: Exact diffusion inversion via coupled transformations},
  author={Wallace, Bram and Gokul, Akash and Naik, Nikhil},
  booktitle={CVPR},
  year={2023}
}

@inproceedings{jupnp,
  title={Pn{P} {I}nversion: Boosting diffusion-based editing with 3 lines of code},
  author={Ju, Xuan and Zeng, Ailing and Bian, Yuxuan and Liu, Shaoteng and Xu, Qiang},
  booktitle={ICLR},
  year={2023}
}

@inproceedings{pan2023effective,
  title={Effective real image editing with accelerated iterative diffusion inversion},
  author={Pan, Zhihong and Gherardi, Riccardo and Xie, Xiufeng and Huang, Stephen},
  booktitle={ICCV},
  year={2023}
}

@inproceedings{garibi2024renoise,
  title={Re{N}oise: Real image inversion through iterative noising},
  author={Garibi, Daniel and Patashnik, Or and Voynov, Andrey and Averbuch-Elor, Hadar and Cohen-Or, Daniel},
  booktitle={ECCV},
  
  year={2024}
}

@inproceedings{hong2024exact,
  title={On exact inversion of {DPM}-{S}olvers},
  author={Hong, Seongmin and Lee, Kyeonghyun and Jeon, Suh Yoon and Bae, Hyewon and Chun, Se Young},
  booktitle={CVPR},
  year={2024}
}

@inproceedings{zhang2024exact,
  title={Exact diffusion inversion via bidirectional integration approximation},
  author={Zhang, Guoqiang and Lewis, Jonathan P and Kleijn, W Bastiaan},
  booktitle={ECCV},
  year={2024}
}

@inproceedings{brack2024ledits++,
  title={L{EDITS}++: Limitless image editing using text-to-image models},
  author={Brack, Manuel and Friedrich, Felix and Kornmeier, Katharia and Tsaban, Linoy and Schramowski, Patrick and Kersting, Kristian and Passos, Apolin{\'a}rio},
  booktitle={CVPR},
  year={2024}
}

@inproceedings{feng2025dit4edit,
  title={Di{T}4{E}dit: {D}iffusion transformer for image editing},
  author={Feng, Kunyu and Ma, Yue and Wang, Bingyuan and Qi, Chenyang and Chen, Haozhe and Chen, Qifeng and Wang, Zeyu},
  booktitle={AAAI},
  year={2025}
}

@inproceedings{miyake2025negative,
  title={Negative-prompt inversion: Fast image inversion for editing with text-guided diffusion models},
  author={Miyake, Daiki and Iohara, Akihiro and Saito, Yu and Tanaka, Toshiyuki},
  booktitle={WACV},
  year={2025},

}

@inproceedings{han2024proxedit,
  title={Prox{E}dit: Improving tuning-free real image editing with proximal guidance},
  author={Han, Ligong and Wen, Song and Chen, Qi and Zhang, Zhixing and Song, Kunpeng and Ren, Mengwei and Gao, Ruijiang and Stathopoulos, Anastasis and He, Xiaoxiao and Chen, Yuxiao and others},
  booktitle={WACV},
  year={2024}
}

@inproceedings{zhao2023null,
  title={Null-text guidance in diffusion models is secretly a cartoon-style creator},
  author={Zhao, Jing and Zheng, Heliang and Wang, Chaoyue and Lan, Long and Huang, Wanrong and Yang, Wenjing},
  booktitle={ACM MM},
  year={2023}
}

@inproceedings{wang2025taming,
  title     = {Taming rectified flow for inversion and editing},
  author    = {Wang, Jiangshan and Pu, Junfu and Qi, Zhongang and Guo, Jiayi and Ma, Yue and Huang, Nisha and Chen, Yuxin and Li, Xiu and Shan, Ying},
  booktitle = {ICML},
  year      = {2025}
}

@inproceedings{
deng2025fireflow,
title={Fire{F}low: Fast inversion of Rectified Flow for Image Semantic Editing},
author={Yingying Deng and Xiangyu He and Changwang Mei and Peisong Wang and Fan Tang},
booktitle={ICML},
year={2025}
}

@article{ma2025adams,
  title={Adams {B}ashforth {M}oulton Solver for Inversion and Editing in Rectified Flow},
  author={Ma, Yongjia and Di, Donglin and Liu, Xuan and Chen, Xiaokai and Fan, Lei and Su, Tonghua and Gao, Yue},
  journal={arXiv preprint arXiv:2503.16522},
  year={2025}
}

@article{mao2025tweezeedit,
  title={Tweeze{E}dit: Consistent and Efficient Image Editing with Path Regularization},
  author={Mao, Jianda and Wang, Kaibo and Xiang, Yang and Chen, Kani},
  journal={arXiv preprint arXiv:2508.10498},
  year={2025}
}

@inproceedings{kulikov2024flowedit,
  title={Flow{E}dit: Inversion-free text-based editing using pre-trained flow models},
  author={Kulikov, Vladimir and Kleiner, Matan and Huberman-Spiegelglas, Inbar and Michaeli, Tomer},
  booktitle={ICCV},
  year={2025}
}

@article{beaudouin2025delta,
  title={Delta Velocity Rectified Flow for Text-to-Image Editing},
  author={Beaudouin, Gaspard and Li, Minghan and Kim, Jaeyeon and Yoon, Sunghoon and Wang, Mengyu},
  journal={arXiv preprint arXiv:2509.05342},
  year={2025}
}

@article{kim2025flowalign,
  title={{F}low{A}lign: {T}rajectory-regularized, inversion-free flow-based image editing},
  author={Kim, Jeongsol and Hong, Yeobin and Park, Jonghyun and Ye, Jong Chul},
  journal={arXiv preprint arXiv:2505.23145},
  year={2025}
}

@inproceedings{xie2025dnaedit,
  title={{DNAE}dit: {D}irect Noise Alignment for Text-Guided Rectified Flow Editing},
  author={Xie, Chenxi and Li, Minghan and Li, Shuai and Wu, Yuhui and Yi, Qiaosi and Zhang, Lei},
  booktitle={NeurIPS},
  year={2025}
}

@article{liu2022rectified,
  title={Rectified flow: A marginal preserving approach to optimal transport},
  author={Liu, Qiang},
  journal={arXiv preprint arXiv:2209.14577},
  year={2022}
}

@inproceedings{lipmanflow,
  title={Flow Matching for Generative Modeling},
  author={Lipman, Yaron and Chen, Ricky TQ and Ben-Hamu, Heli and Nickel, Maximilian and Le, Matthew},
  booktitle={ICLR},
 year={2023}
}

@article{huynh2008scope,
  title={Scope of validity of {PSNR} in image/video quality assessment},
  author={Huynh-Thu, Quan and Ghanbari, Mohammed},
  journal={Electronics {L}etters},
  
  year={2008},
  publisher={IET}
}

@article{wang2004image,
  title={Image quality assessment: from error visibility to structural similarity},
  author={Wang, Zhou and Bovik, Alan C and Sheikh, Hamid R and Simoncelli, Eero P},
  journal={IEEE Transactions on Image Processing},
  year={2004}
}

@inproceedings{zhang2018unreasonable,
  title={The Unreasonable Effectiveness of Deep Features as a Perceptual Metric},
  author={Zhang, Richard and Isola, Phillip and Efros, Alexei A and Shechtman, Eli and Wang, Oliver},
  booktitle={CVPR},
  year={2018}
}

@inproceedings{radford2021learning,
  title={Learning transferable visual models from natural language supervision},
  author={Radford, Alec and Kim, Jong Wook and Hallacy, Chris and Ramesh, Aditya and Goh, Gabriel and Agarwal, Sandhini and Sastry, Girish and Askell, Amanda and Mishkin, Pamela and Clark, Jack and others},
  booktitle={ICML},
  
  year={2021},

}

@inproceedings{caron2021emerging,
  title={Emerging properties in self-supervised vision transformers},
  author={Caron, Mathilde and Touvron, Hugo and Misra, Ishan and J{\'e}gou, Herv{\'e} and Mairal, Julien and Bojanowski, Piotr and Joulin, Armand},
  booktitle={ICCV},

  year={2021}
}

@inproceedings{song2020denoising,
  title={Denoising diffusion Implicit Models},
  author={Song, Jiaming and Meng, Chenlin and Ermon, Stefano},
  booktitle={ICLR},
  year={2021}
}

@inproceedings{ho2020denoising,
  title={Denoising diffusion probabilistic models},
  author={Ho, Jonathan and Jain, Ajay and Abbeel, Pieter},
  booktitle={NeurIPS},

  year={2020}
}

@article{song2020score,
  title={Score-based generative modeling through stochastic differential equations},
  author={Song, Yang and Sohl-Dickstein, Jascha and Kingma, Diederik P and Kumar, Abhishek and Ermon, Stefano and Poole, Ben},
  journal={ICLR},
  year={2021}
}

@article{li2025exact,
  title={On Exact Editing of Flow-Based Diffusion Models},
  author={Li, Zixiang and Song, Yue and Peng, Jianing and Liu, Ting and Huang, Jun and Qu, Xiaochao and Liu, Luoqi and Wang, Wei and Zhao, Yao and Wei, Yunchao},
  journal={arXiv preprint arXiv:2512.24015},
  year={2025}
}

@inproceedings{
zhao2024wedit,
title={W-{EDIT}: A Wavelet-Based Frequency-Aware Framework for Text-Driven Image Editing},
author={Jiahui Sun and Weining Wang and Mingzhen Sun and Peiyao Wang and Xinxin Zhu and Jing Liu},
booktitle={ICLR},
year={2026},

}

@inproceedings{wu2024freediff,
  title={Free{D}iff: Progressive frequency truncation for image editing with diffusion models},
  author={Wu, Wei and Fan, Qingnan and Qin, Shuai and Gu, Hong and Zhao, Ruoyu and Chan, Antoni B},
  booktitle={ECCV},

  year={2024}
}

@inproceedings{teng2024fsi,
  title={{FSI-Edit}: Frequency and Stochasticity Injection for Flexible Diffusion-Based Image Editing},
  author={Yang, Kaixiang and Li, Xin and Li, Yuxi and Li, Qiang and Wang, Zhiwei},
  booktitle={NeurIPS},
  year={2025}
}

@inproceedings{tumanyan2023plug,
  title={Plug-and-play diffusion features for text-driven image-to-image translation},
  author={Tumanyan, Narek and Geyer, Michal and Bagon, Shai and Dekel, Tali},
  booktitle={CVPR},
  year={2023}
}

@inproceedings{cao2023masactrl,
  title     = {{MasaCtrl}: Tuning-Free Mutual Self-Attention Control for Consistent Image Synthesis and Editing},
  author    = {Cao, Mingdeng and Wang, Xintao and Qi, Zhongang and Shan, Ying and Qie, Xiaohu and Zheng, Yinqiang},
  booktitle = {ICCV},
  year      = {2023}
}

@inproceedings{liu2024fia,
  title={{FIA-Edit}: {F}requency-Interactive Attention for Efficient and High-Fidelity Inversion-Free Text-Guided Image Editing},
  author={Yang, Kaixiang and Shen, Boyang and Li, Xin and Dai, Yuchen and Luo, Yuxuan and Ma, Yueran and Fang, Wei and Li, Qiang and Wang, Zhiwei},
  booktitle={AAAI},
  year={2026}
}

@inproceedings{hertz2022prompt,
  title={Prompt-to-{P}rompt Image Editing with Cross Attention Control},
  author={Hertz, Amir and Mokady, Ron and Tenenbaum, Jay and Aberman, Kfir and Pritch, Yael and Cohen-Or, Daniel},
  booktitle={ICLR},
  year={2023}
}

@inproceedings{xu2024inversion,
  title={Inversion-free image editing with language-guided diffusion models},
  author={Xu, Sihan and Huang, Yidong and Pan, Jiayi and Ma, Ziqiao and Chai, Joyce},
  booktitle={CVPR},

  year={2024}
}

@inproceedings{rout2025semantic,
  title={Semantic Image Inversion and Editing using Rectified Stochastic Differential Equations},
  author={Rout, L and Chen, Y and Ruiz, N and Caramanis, C and Shakkottai, S and Chu, W},
  booktitle={ICLR},
  year={2025}
}

@inproceedings{avrahami2025stable,
  title={Stable {F}low: Vital layers for training-free image editing},
  author={Avrahami, Omri and Patashnik, Or and Fried, Ohad and Nemchinov, Egor and Aberman, Kfir and Lischinski, Dani and Cohen-Or, Daniel},
  booktitle={CVPR},

  year={2025}
}

@inproceedings{esser2024scaling,
  title={Scaling rectified flow transformers for high-resolution image synthesis},
  author={Esser, Patrick and Kulal, Sumith and Blattmann, Andreas and Entezari, Rahim and M{\"u}ller, Jonas and Saini, Harry and Levi, Yam and Lorenz, Dominik and Sauer, Axel and Boesel, Frederic and others},
  booktitle={ICML},
  year={2024}
}

@misc{flux2024,
    author={Black Forest Labs},
    title={FLUX},
    year={2024},
    howpublished={\url{https://github.com/black-forest-labs/flux}},
}

@inproceedings{
martin2025pnpflow,
title={{PnP-F}low: {P}lug-and-Play Image Restoration with Flow Matching},
author={S{\'e}gol{\`e}ne Tiffany Martin and Anne Gagneux and Paul Hagemann and Gabriele Steidl},
booktitle={ICLR},
year={2025}
}

@article{mallat2002theory,
  title={A theory for multiresolution signal decomposition: {T}he wavelet representation},
  author={Mallat, Stephane G},
  journal={IEEE Transactions on Pattern Analysis and Machine Intelligence},  
  year={2002},
  publisher={IEEE}
}

@inproceedings{kim2022diffusionclip,
  title={Diffusion{CLIP}: Text-guided diffusion models for robust image manipulation},
  author={Kim, Gwanghyun and Kwon, Taesung and Ye, Jong Chul},
  booktitle={CVPR},
  year={2022}
}

@article{cooley1965algorithm,
  title={An algorithm for the machine calculation of complex Fourier series},
  author={Cooley, James W and Tukey, John W},
  journal={Mathematics of {C}omputation},

  year={1965},
  publisher={JSTOR}
}

@article{butterworth1930theory,
  title={On the theory of filter amplifiers},
  author={Butterworth, Stephen and others},
  journal={Wireless {E}ngineer},

  year={1930}
}

@article{tschirsich2015notes,
  title={Notes on discrete Gaussian scale space},
  author={Tschirsich, Martin and Kuijper, Arjan},
  journal={Journal of {M}athematical {I}maging and {V}ision},
 
  year={2015},

}

@book{jahne2005digital,
  title={Digital image processing},
  author={J{\"a}hne, Bernd},
  year={2005},
  publisher={Springer}
}

@inproceedings{lin2024schedule,
  title={Schedule your edit: {A} simple yet effective diffusion noise schedule for image editing},
  author={Lin, Haonan and Chen, Yan and Wang, Jiahao and An, Wenbin and Wang, Mengmeng and Tian, Feng and Liu, Yong and Dai, Guang and Wang, Jingdong and Wang, Qianying},
  booktitle={NeurIPS},
  year={2024}
}
